%% file: acl_latex.tex
\documentclass[11pt]{article}

\usepackage{acl}

\usepackage{times}
\usepackage{latexsym}

\usepackage[T1]{fontenc}
\usepackage[utf8]{inputenc}

\usepackage{microtype}

\usepackage{inconsolata}

\usepackage{graphicx}

\usepackage{array}
\usepackage{ragged2e}
\usepackage{booktabs}
\usepackage{multirow}
\usepackage{makecell}
\usepackage{adjustbox}
\usepackage{amsmath}
\newcommand{\brk}[1]{\par\noindent(#1)}
\usepackage{xcolor}
\usepackage[most]{tcolorbox}
\usepackage{listings}
\usepackage{float}
\usepackage{cuted}
\usepackage{tabularx}   % auto-width column tables
\usepackage{amssymb}   % star symbols etc.
\usepackage[linesnumbered,ruled,vlined]{algorithm2e} 
\SetKw{Continue}{continue}
\SetKw{Break}{break}
\DontPrintSemicolon
\title{LLM Benchmark--User Need Misalignment for Climate Change}
\author{
  Oucheng Liu,
  Lexing Xie
  \and
  Jing Jiang
  \\
  \ \\
  School of Computing
  \\
  The Australian National University
  \\
  Canberra, Australia
  \\
  \texttt{\{oucheng.liu,lexing.xie,jing.jiang\}@anu.edu.au}
}

\begin{document}
\maketitle
\begin{abstract}
Climate change is a major socio-scientific issue shapes public decision-making and policy discussions. As large language models (LLMs) increasingly serve as an interface for accessing climate knowledge, whether existing benchmarks reflect user needs is critical for evaluating LLM in real-world settings. We propose a Proactive Knowledge Behaviors Framework that captures the different human–human and human–AI knowledge seeking and provision behaviors. We further develop a Topic--Intent--Form taxonomy and apply it to analyze climate-related data representing different knowledge behaviors. Our results reveal a substantial mismatch between current benchmarks and real-world user needs, while knowledge interaction patterns between humans and LLMs closely resemble those in human--human interactions. These findings provide actionable guidance for benchmark design, RAG system development, and LLM training. Code is available at \url{https://github.com/OuchengLiu/LLM-Misalign-Climate-Change}.
\end{abstract}

\input{Sections/1_Introduction}
\input{Sections/2_Related_Work}
\input{Sections/3_Framework}

\input{Sections/4_Methods}

\input{Sections/5_Findings}

\input{Sections/6_Conclusion}
\input{Sections/7_Limitation}

\bibliography{custom}

\appendix
\input{Sections/Appendix_1}
\input{Sections/Appendix_2}

\input{Sections/Appendix_3}

\end{document}

%% file: Sections/1_Introduction.tex
\section{Introduction}
\label{sec:intro}

Climate change is a critical socio-scientific challenge whose impacts extend beyond climate science to domains such as food systems, public health, and economic development~\citep{IPCC_AR6_WG1_2021,IPCC_AR6_WG2_2022,IPCC_AR6_WG3_2022}. As climate risks increasingly affect societies, public demand for climate-related knowledge continues to grow~\citep{ClimateQ&A}. At the same time, large language models (LLMs) are rapidly becoming common interfaces through which people seek information and produce written content~\citep{HowPeopleUseChatGPT}. This raises concerns about the reliability of AI-generated climate information, particularly when non-expert users may over-trust and further disseminate inaccurate or misleading outputs~\citep{RiskMisinformation,LLMHallucinationSpreading,LLMSocialInfluences}. These concerns highlight the importance of evaluation benchmarks that reflect real-world user needs.

However, it remains unclear whether existing benchmarks used to evaluate LLM knowledge of climate change truly reflect the questions that users ask when consulting LLMs about climate change. In particular, it is uncertain whether these benchmarks accurately capture the diversity of topics, user intents, and expected answer forms that arise in real-world interactions.

To address this question, we first perform a systematic comparison between datasets representing real-world needs and existing benchmarks. Specifically, we identify climate-change–related queries from real user–LLM interaction datasets, including \textit{WildChat}~\citep{WildChat}, \textit{LMSYS-Chat}~\citep{LMSYS_Chat_1M}, and \textit{ClimateQ\&A}~\citep{ClimateQ&A}. We also design an LLM-based topic modeling approach and construct taxonomies of topics, intents, and answer forms to annotate the data. Our results reveal a significant misalignment. Existing benchmarks focus on a narrow subset of climate knowledge and limited question types, whereas real user needs cover a broader range of topics (e.g., policy, transition and action), require higher-level procedural and metacognitive support (e.g., advice and actionable writing), and often demand more structured output formats (e.g., explanatory paragraphs or itemized lists).

\begin{figure}[t]
    \centering
    \includegraphics[width=\columnwidth, page=1]{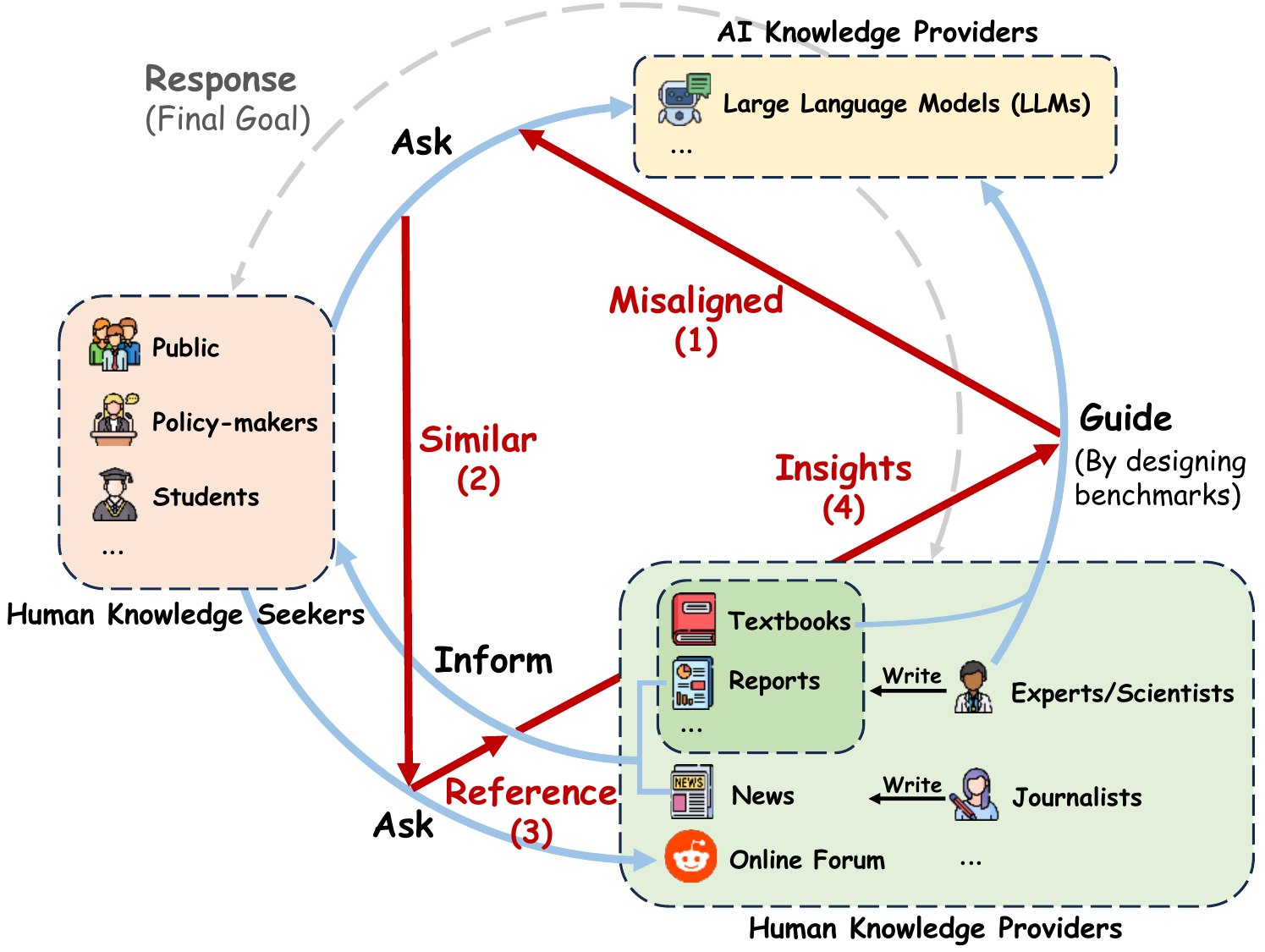}
    \caption{Our Proactive Knowledge Behaviors Framework. The proactive knowledge behaviors between three key actors are shown as blue arrows and the red arrows reflect our analytical logic.}
    \label{fig:framework}
\end{figure}

A natural solution is to construct benchmarks directly from user queries. However, such queries are typically open-ended, diverse, and context-dependent, which makes it difficult to obtain reference answers at scale. This raises an important question: are there other readily accessible sources of high-quality knowledge that closely match these real-world knowledge needs? Here, we propose a \textit{Proactive Knowledge Behaviors Framework} (see Figure~\ref{fig:framework} and Section~\ref{sec:framework}) to conceptualize human \textit{knowledge seeking} (i.e., asking) and \textit{knowledge provision} (i.e., guiding, informing) behaviors. The framework serves to guide the empirical discovery and identification of candidate knowledge sources.

Our analysis guided by the framework reveals that \textit{human--human} and \textit{human--AI} knowledge interactions are fundamentally aligned on knowledge needs. Specifically, by comparing the topic--intent--form distributions of questions asked when humans seek knowledge from other humans with those queries posed to LLMs, we observe a strong similarity. This alignment provides the premise for directly leveraging data from \textit{human--human} knowledge interactions as a reference to support \textit{human--AI} knowledge interactions. Furthermore, we find that within human-to-human knowledge provision, the topic distribution of the comprehensive and authoritative \textit{IPCC AR6} reports~\citep{IPCC_AR6_WG1_2021,IPCC_AR6_WG2_2022,IPCC_AR6_WG3_2022} are highly consistent with human needs. This suggests that it can serve as a high-quality minimal knowledge base when constructing benchmarks (through sub-sampling), or when building retrieval-augmented generation (RAG) retrieval corpora. In summary, in this work we make the following contributions:
\begin{itemize}
\item We identify a misalignment between existing climate change LLM benchmarks and real-world knowledge needs.
\item We provide reference distributions of topic--intent--form and corpora for climate-change LLM evaluation and development.
\item We introduce reusable taxonomies for climate-change topics, user intents, and answer forms.
% \item \textbf{A transferable methodology for socio-scientific issues}: An analogy-inspired analytical framework for guiding LLM evaluation and development, along with a practical workflow for large-scale topic modeling.
\end{itemize}

%% file: Sections/2_Related_Work.tex
\section{Related Work}

In the climate change domain, most previous NLP resources are designed for well-defined traditional tasks such as stance detection~\citep{ClimateStance,lobbyMap}, fact verification~\citep{Climate_FEVER}, claims identification~\citep{Environmental_Claims} and information retrieval~\citep{climRetrieve}, which are significantly different from open-ended human-LLM interactions. Recent work has begun to explore LLM-based climate question answering and communication, including literacy evaluation~\citep{AccessClimate}, QA benchmarks~\citep{ClimaQA}, grounded dialogue systems~\citep{ChatClimate,climateGPT,chatNetZero}, multi-source data integration~\citep{MultisourceLLM}, and personalized recommendations~\citep{Climate_Advisor}. Nevertheless, these efforts remain largely science-oriented, leaving open questions about how to better support real-world, user-facing climate-related LLM interactions.

%% file: Sections/3_Framework.tex
\section{Framework and Data}
\label{sec:framework}

\begin{table*}[htbp]
\centering
\begin{adjustbox}{max width=\textwidth}
\scriptsize
\setlength{\tabcolsep}{4pt}
\renewcommand{\arraystretch}{1.15}
\begingroup
\setlength{\parindent}{0pt}
\setlength{\aboverulesep}{0.6ex}
\setlength{\belowrulesep}{0.6ex}
\setlength{\cmidrulesep}{0.4ex}
\begin{tabular}{
  >{\RaggedRight\arraybackslash}p{0.15\linewidth}  % Category
  >{\RaggedRight\arraybackslash}p{0.17\linewidth}  % Dataset
  >{\RaggedRight\arraybackslash}p{0.20\linewidth}  % Task/Format
  >{\RaggedRight\arraybackslash}p{0.41\linewidth}  % Brief Description
  >{\RaggedRight\arraybackslash}p{0.07\linewidth}   % Count
}
\toprule
\textbf{Category} & \textbf{Dataset} & \textbf{Task/Format} & \textbf{Brief Description} & \textbf{Count} \\
\midrule

% Category 1
\multirow{2}{*}{\makecell[l]{\textit{Human-to-AI}\\\textit{Queries}}} &
WildChat &
LLM--User Conversations (Queries) &
Large-scale multilingual LLM--user logs; diverse IPs/languages/regions/domains. & 1,706 \\
\cmidrule[0.6pt](lr){2-5}
& LMSYS-Chat-1M &
LLM--User Conversations (Queries) &
Large-scale multilingual LLM--user logs; diverse IPs/languages/regions/domains. & 1,331 \\
\cmidrule[0.6pt](lr){2-5}
& ClimateQ\&A &
LLM--User Conversations (Queries) &
Authentic climate-related questions from French public to a climate-specialized LLM. & 3,033 \\
\midrule

% Category 3
\multirow{1}{*}{\makecell[l]{\textit{Human-to-Human}\\\textit{Questions}}} &
Reddit \brk{4 subreddits} &
Questions \brk{Forum Posts} &
Questions from \textit{r/climatechange}, \textit{r/climate}, \textit{r/climate\_science}, \textit{r/GlobalClimateChange}; collected from top/new/hot, deduplicated, interrogatives only. & 1,023 \\
\midrule

% Category 2
\multirow{2}{*}{\makecell[l]{\textit{Human-to-AI}\\\textit{Guidance Knowledge}}} &
ClimaQA-Gold &
QA Benchmark \brk{Expert-annotated} &
Graduate-level climate science QA; expert annotated for high reliability/authority. & 540 \\
\cmidrule[0.6pt](lr){2-5}
& ClimaQA-Silver &
QA Benchmark \brk{Synthetic} &
Large-scale synthetic graduate-level climate science QA for broad coverage in evaluation/fine-tuning. & 2,543 \\
\midrule

% Category 4
\multirow{2}{*}{\makecell[l]{\textit{Human-to-Human}\\\textit{Knowledge Provision}}} &
SciDCC &
News Articles &
Climate-related news articles from \textit{Science Daily}; media/journalistic narratives of climate issues. & 7,476 \\
\cmidrule[0.6pt](lr){2-5}
& IPCC \textit{AR6} \brk{WG I/II/III} &
Scientific Reports \brk{Paragraph-level} &
Authoritative synthesis for policymakers/public. WG I: Physical Basis; WG II: Impacts/Adaptation/Vulnerability; WG III: Mitigation. Paragraphs treated as instances. & 18,889 \\
\bottomrule
\end{tabular}
\endgroup
\end{adjustbox}

\caption{An overview of the eight core climate-related datasets used in this study. ``Count'' refers to the number of valid samples after cleaning and filtering. See three ``Auxiliary Corpora'' in Appendix~\ref{app:auxillary}.}
\label{tab:datasets}
\end{table*}

% In this section, we present a framework to conceptually distinguish different knowledge behaviors related to climate change and use this framework to guide data collection for our study. 
Directly searching for solutions from real user queries is costly. When it is not feasible to start from the target scenario, a practical strategy is to draw inspiration from a related and mature pattern. Our \textit{Proactive Knowledge Behaviors Framework} (Figure~\ref{fig:framework}) is built on this idea. Specifically, we leverage mature human--human knowledge interaction patterns to analogy-inspire emerging human--LLM knowledge interactions. Within this framework, we identify three types of actors: human knowledge seekers (e.g., the public, policy makers, and students), human knowledge providers (e.g., experts, journalists, scientists, and contributors on online Q\&A platforms such as Reddit), and AI knowledge providers (i.e., large language models in this work). We focus on proactive knowledge behaviors in both human--human and human--AI interactions. These behaviors fall into two categories: \textit{knowledge seeking} and \textit{knowledge provision}. 

For knowledge seeking, we examine two types of behaviors. The first is human knowledge seekers \textbf{asking} AI knowledge providers. We collect the corresponding data as \textbf{Human-to-AI Queries}. The data come from three sources: \textit{WildChat}~\cite{WildChat} and \textit{LMSYS-Chat-1M}~\citep{LMSYS_Chat_1M}, two public LLM conversation logs, and \textit{ClimateQ\&A}~\citep{ClimateQ&A}, a dataset of climate-related questions posed by the French public to an LLM-based system. For \textit{WildChat} and \textit{LMSYS-Chat-1M}, we use an LLM to filter out conversations unrelated to climate change. From each conversation, we retain only the first-turn user query to capture the user's original intention.

The second type is human knowledge seekers \textbf{asking} human knowledge providers. We define the corresponding data as \textbf{Human-to-Human Questions}, which are collected from question-style posts in four major climate-related subreddits on Reddit~\citep{RedditAPI}.\footnote{The Reddit data used in this study was extracted on 2025-08-04 20:14 AEST (UTC+10:00) via the official Reddit API.}

For knowledge provision, we also consider two forms of behaviors. The first refers to human knowledge providers \textbf{guiding} AI knowledge providers. This refers to guiding LLM development through benchmark design. We collect the corresponding data as \textbf{Human-to-AI Guidance Knowledge} using the \textit{ClimaQA} dataset~\citep{ClimaQA}, one of the most comprehensive benchmarks for evaluating LLM climate knowledge. We analyze its two subsets separately: \textit{ClimaQA-Gold} and \textit{ClimaQA-Silver}. Both subsets were generated by LLMs based on textbooks and include multiple-choice, cloze, and free-form questions. In our analysis, we use only the question components.

The second form is human knowledge providers proactively \textbf{providing} knowledge to human knowledge seekers. We collect the data as \textbf{Human-to-Human Knowledge Provision}. The data come from: \textit{SciDCC}~\citep{SciDCC}, a climate news dataset, and \textit{IPCC AR6 (WG I/II/III)}~\citep{IPCC_AR6_WG1_2021,IPCC_AR6_WG2_2022,IPCC_AR6_WG3_2022}, a highly authoritative and comprehensive assessment report on climate change.

In addition, we collect several auxiliary corpora to support topic modeling. These include \textit{Climate-FEVER}~\citep{Climate_FEVER}, \textit{Environmental Claims}~\citep{Environmental_Claims}, and \textit{ClimSight}~\citep{ClimSight}. Details of all eleven datasets are summarized in Table~\ref{tab:datasets}, with examples provided in Appendix~\ref{app:examples}.

We focus exclusively on proactive behaviors and exclude passive behaviors such as responses, as proactive behaviors determine whether and how knowledge flows emerge and therefore reveal the underlying motivations and information needs in knowledge interactions. Using this framework, we first identify the misalignment between evaluation and demand in human--AI knowledge interactions. We then examine whether knowledge needs in human--AI interactions align with those in human--human interactions. Based on this premise, we discover high-quality data from human--human knowledge interactions and translate them into insights for LLM benchmark construction and LLM development, with the ultimate goal of improving LLM responses on climate change topics.

%% file: Sections/4_Methods.tex
\section{Methods}
\label{sec:methods}

\begin{figure*}[t]
    \centering
    \includegraphics[width=0.98\textwidth, page=1]{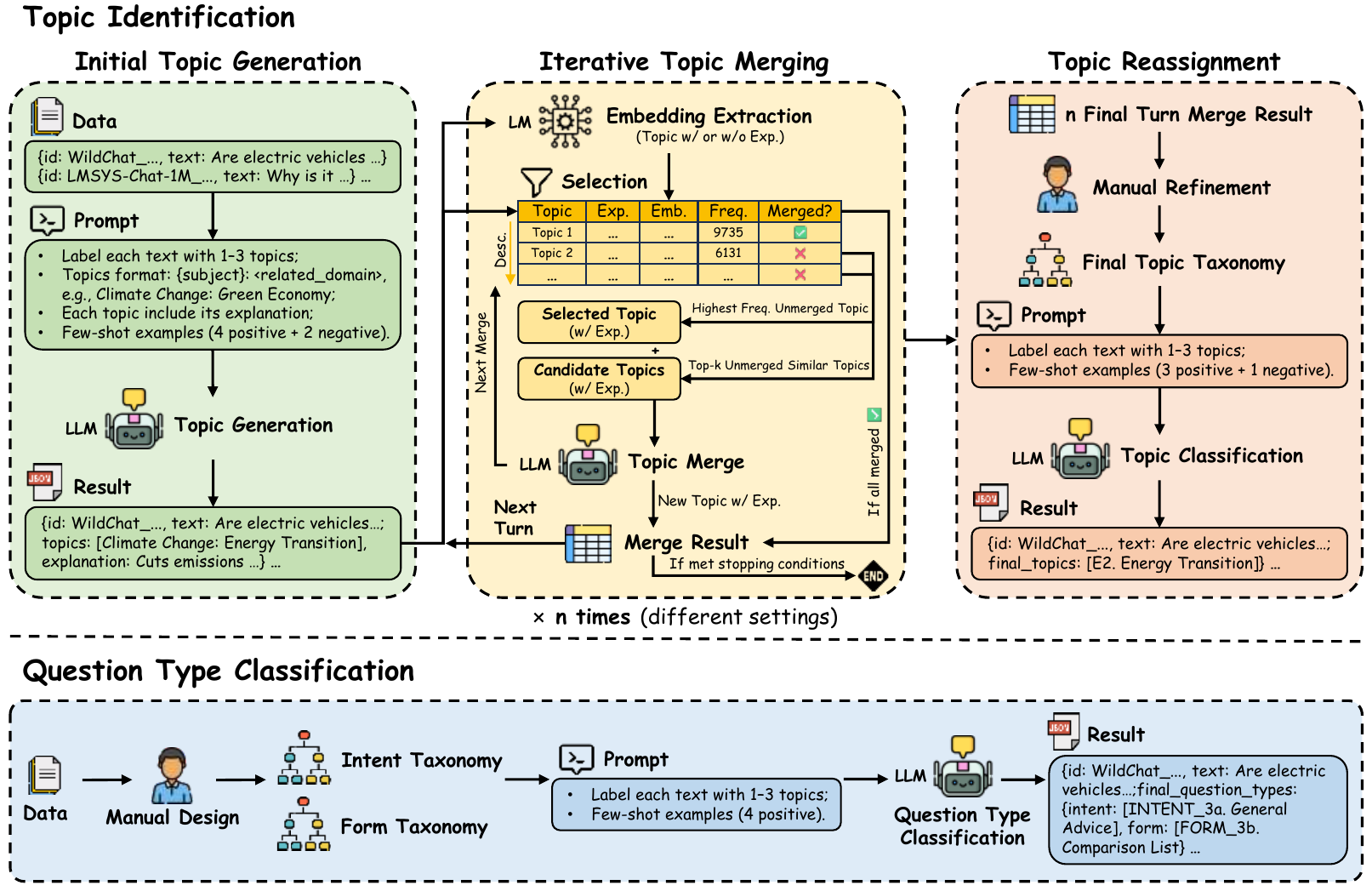}
    \caption{Pipelines for data annotation in Topic Identification and Question Type Classification.}
    \label{fig:annoation}
\end{figure*}

This section describes our methodology for analyzing data. We first construct a topic taxonomy with the help of an LLM. We then further classify each data point along two dimensions—user intent and expected answer form—again leveraging an LLM. Finally, we represent queries as weighted vectors over topics, intents, and forms to enable quantitative analyses.

\subsection{Topic Identification}
\label{sec:topic_indentification}
\paragraph{Initial Topic Generation.}
Traditional topic models such as LDA~\citep{LDA} and BERTopic~\citep{BerTopic}, which rely on lexical co-occurrence or static embeddings, often struggle to accurately capture semantic relations in complex or multi-faceted text. Inspired by TopicGPT and HICode~\citep{TopicGPT,HICode}, we instead leverage the stronger semantic reasoning capabilities of LLMs to generate free-form topics. For each data sample, we apply a 6-shot in-context learning prompt (see Appendix~\ref{app:Prompt 1}) to instruct the LLM to generate one to three topic labels together with brief explanations. Samples judged by the model as irrelevant to climate change are removed. 
Details are provided in Appendix~\ref{app:LLMuse}.

\paragraph{Iterative Topic Merging.}
Free-form topic generation often yields redundant or semantically overlapping labels, necessitating topic merging. We sort topics by descending frequency and compute LM-based embeddings. At each step, the most frequent unmerged topic is used as an anchor to retrieve its top-$k$ most similar topics by cosine similarity. An LLM then merges only semantically equivalent or granularity-reconcilable topics into a new topic and description (Appendix~\ref{app:Prompt 2}). One turn of merging processes all topics once, and we iteratively run multiple turns of merging until no merging occurs. Finally, the topic set is reduced to 1.58\%--6.73\% of its original size.

To improve coverage and robustness, we repeat the merging process under multiple settings (Appendix~\ref{app:topic_merge_settings}): varying the use of explanations, embedding LMs, and merging LLMs. These variations mitigate missed merges due to embedding limitations and account for the inherently subjective nature of taxonomy construction, reducing bias introduced by any single configuration.

\paragraph{Topic Reassignment.}
Although a single round of merging already produces a high-quality topic list, we further conduct manual refinement across multiple merging results to construct a comprehensive and well-defined final taxonomy. For each merge setting, we take the union of the top-5 most frequent topics across the eleven datasets and supplement essential top-10 topics (Appendix~\ref{app:manual_topic_reassignment_details}). This process standardizes topic granularity and merges residual semantic overlaps. The final taxonomy adopts a two-level hierarchy (Figure~\ref{fig:topic_taxonomy} in Appendix~\ref{app:Taxonomies}), consisting of five coarse-grained categories (i.e., \textit{Climate Science Foundations \& Method}, \textit{Ecological Impacts}, \textit{Human Systems \& Socioeconomic Impacts}, \textit{Adaptation Strategies}, and \textit{Mitigation Mechanisms}, twenty-five fine-grained topics (e.g., \textit{Biodiversity Loss}, \textit{Agriculture \& Food Security}, and \textit{Energy Transition}), and an \textit{Others} category. After establishing the taxonomy, we reassign topics to samples using an LLM with 4-shot in-context learning, allowing each sample to be associated with up to three relevant topics to capture multi-topic or coupled cases.

\subsection{Question Type Classification}

We analyze queries' question type through a taxonomy defined by two dimensions: \textbf{user intent} and \textbf{expected answer form}, capturing not only what users ask but also what they expect in response. Unlike topic modeling, both dimensions admit relatively clear ground-truth labels, enabling a fixed taxonomy design (Figure~\ref{fig:question_type} in Appendix~\ref{app:Taxonomies}). Although our analysis focuses on climate change, the taxonomy is intentionally designed to be broadly applicable.
% as queries within a domain cannot be assumed to follow a small, predefined set of intents or forms. 
To ensure a comprehensive coverage and robustness to evolving LLM capabilities, our design builds on prior work~\citep{ShiftedOverlooked,URS,TnT-LLM,Microsoft_User_Intent} and incorporates empirical observations from our data as well as newly introduced commercial LLM functionalities~\citep{OenaAI_Agent}.

User intent and answer form are each organized into eight major categories, comprising twenty-nine types in total, along with nine ``Others'' types. For instance, intent includes \textit{Fact Lookup} and \textit{General Advice}, while form includes \textit{Concise Paragraph} and \textit{Item List}.  Each intent type is further associated with a \textbf{knowledge-type label} derived from an extended Bloom’s taxonomy~\citep{Knowledge_Taxonomy}, including \textit{Factual}, \textit{Conceptual}, \textit{Procedural}, and \textit{Metacognitive} knowledge (Appendix~\ref{app:Taxonomies}). This mapping enables us to infer both the question category and the cognitive competencies required of the LLM. Finally, we use an LLM to label each instance with between 1 and 3 ranked intent and form labels. Details regarding the human verification for LLM annotation can be found in Appendix~\ref{app:human_verification}.

% \subsection{Human Verification} 
% We recruited six human annotators to manually validate the LLM-annotated data. We randomly sampled 150 data instances from the full dataset, including 15 instances from each of the eight core datasets and 10 instances from each of the three auxiliary datasets. The LLM-annotated data achieved Jaccard scores of 0.706, 0.743, and 0.783 on topic, intent and form, respectively. 

\subsection{Method for Analysis}
For quantitative analysis, we retained the {42{,}261} samples classified under the {25 finalized topics}. Each sample was represented as weighted vectors along three dimensions—Topic, Intent, and Form—where higher-ranked labels received greater weights normalized to a sum of one. Specifically, if a sample in a given dimension has $K$ ordered labels (sorted by relevance), their weights follow the ratio: $w_1 : w_2 : \cdots : w_K=K : (K-1) : \cdots : 1$. All subsequent analyses were based on these weighted representations. To compare distributions across datasets and groups, we computed cosine similarities between normalized fixed-length weight vectors (\(25\) dimensions for Topics and \(38\) for Intents and Forms), quantifying the alignment and divergence of knowledge behaviors across data sources.

%% file: Sections/5_Findings.tex
\section{Results}

\subsection{Benchmark-User Need Misalignment}
\label{r:result1}
\begin{figure}[t]
    \centering
    \includegraphics[width=\columnwidth, page=1]{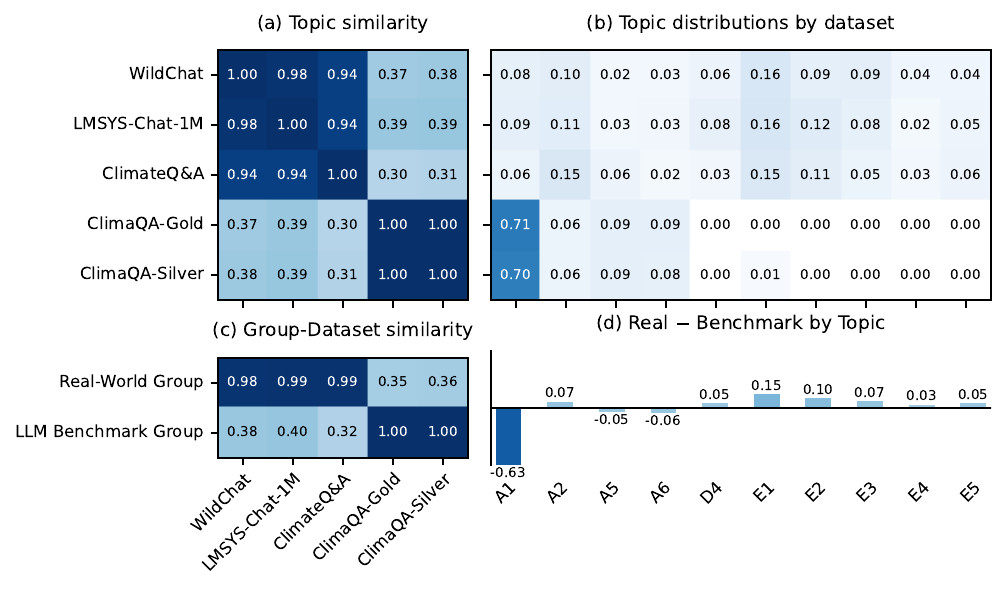}
    \caption{Topic comparison between Human-to-AI Queries and Human-to-AI Guidance Knowledge. (a) Pairwise topic-distribution similarities across the five datasets; (b) Probabilities of the 10 most diverging topics, i.e., those topics with the highest absolute probability differences under the two groups; (c) Topic-distribution similarities between each dataset and each group; (d) Probability differences of the most diverging topics. 
    % Notes: In (b) and (d), only the Top-10 topics by absolute distribution difference are shown. 
    The interpretations and presentations of Figures~\ref{fig:rq1_intent} and~\ref{fig:rq1_form} follow the same logic.}
    \label{fig:rq1_topic}
\end{figure}

To evaluate \textbf{whether current LLM benchmarks for climate change reflect real-world knowledge needs}, we compare Human-to-AI Queries and Human-to-AI Guidance Knowledge. Recall that the former includes \textit{WildChat}, \textit{LMSYS-Chat-1M}, and \textit{ClimateQ\&A} while the latter includes \textit{ClimaQA-Gold} and \textit{ClimaQA-Silver}.
The three Human-to-AI query datasets are merged as the \textbf{Real-World Group}, while the two Human-to-AI Guidance Knowledge datasets are merged as the \textbf{LLM Benchmark Group}, with each dataset weighted proportionally to its size rather than applying uniform weights. 
Figure~\ref{fig:rq1_topic}(a) shows pairwise cosine similarities of topic distributions among the five datasets, while Figure~\ref{fig:rq1_topic}(c) shows cosine similarities between each dataset and the two groups. These results confirm the validity of grouping: the three Human-to-AI query datasets exhibit highly consistent topic distributions (with similarities 0.94–0.98), indicating strong real-world representativeness across distinct data sources, while the two Human-to-AI Guidance Knowledge datasets show nearly identical internal distributions.

We now examine the divergence of the two groups in terms of \textbf{Topic}. 
Figure~\ref{fig:rq1_topic}(b) and \ref{fig:rq1_topic}(d) show that the topic distribution of the Real-World Group is more dispersed, with non-trivial proportions spanning many topics; in contrast, the LLM Benchmark Group is highly concentrated in \textit{A1. Atmospheric Science \& Climate Processes}, where its share exceeds that of the Real-World Group by approximately 63\%. Meanwhile, the Real-World Group shows substantially greater attention to \textit{E. Mitigation Mechanisms}, particularly \textit{E1. Climate Policy, Governance \& Finance Mechanism} and \textit{E2. Energy Transition}, while the LLM Benchmark Group pays almost no attention to these categories.

\begin{figure}[t]
    \centering
    \includegraphics[width=\columnwidth, page=1]{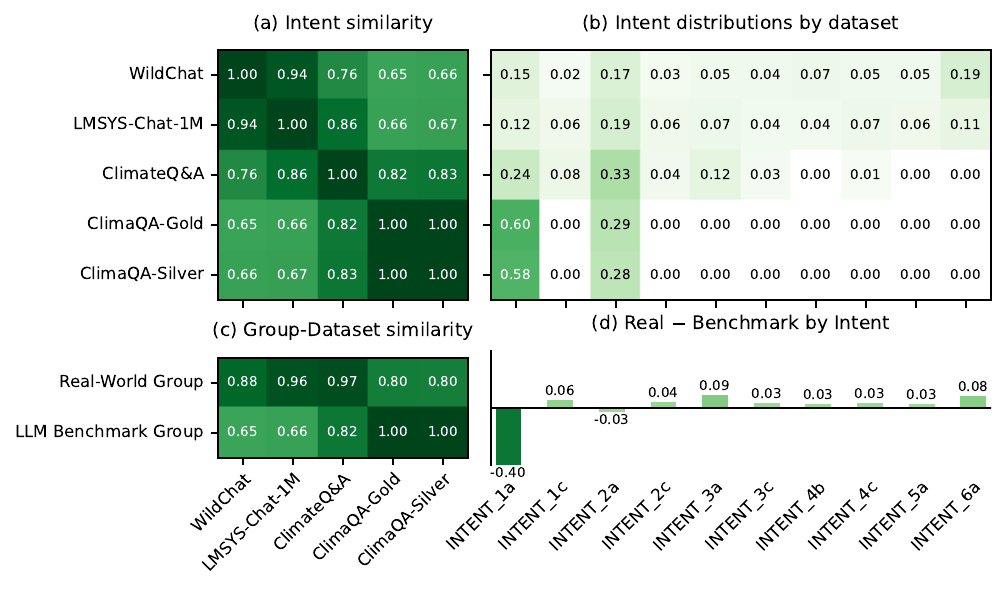}
    \caption{User intents comparison between Human-to-AI Queries and Human-to-AI Guidance Knowledge.}
    \label{fig:rq1_intent}
\end{figure}

For comparison regarding \textbf{User Intent}, which is shown in Figure~\ref{fig:rq1_intent}, all datasets contain non-negligible proportions~($\geq$12\%) of \textit{INTENT\_1a. Fact Lookup} and \textit{INTENT\_2a. Reasoning / Causal Analysis}. The Benchmark Group allocates as much as 60\% to \textit{INTENT\_1a. Fact Lookup}, exceeding the Real-World Group by roughly 40\% on average. In contrast, Figure~\ref{fig:rq1_intent}(b) indicates that real-user intents are more diverse, and Figure~\ref{fig:rq1_intent}(d) shows that humans more frequently seek \textit{INTENT\_3a. General Advice} and \textit{INTENT\_6a. Operational Writing}. 

\begin{figure}[t]
    \centering
    \includegraphics[width=\columnwidth, page=1]{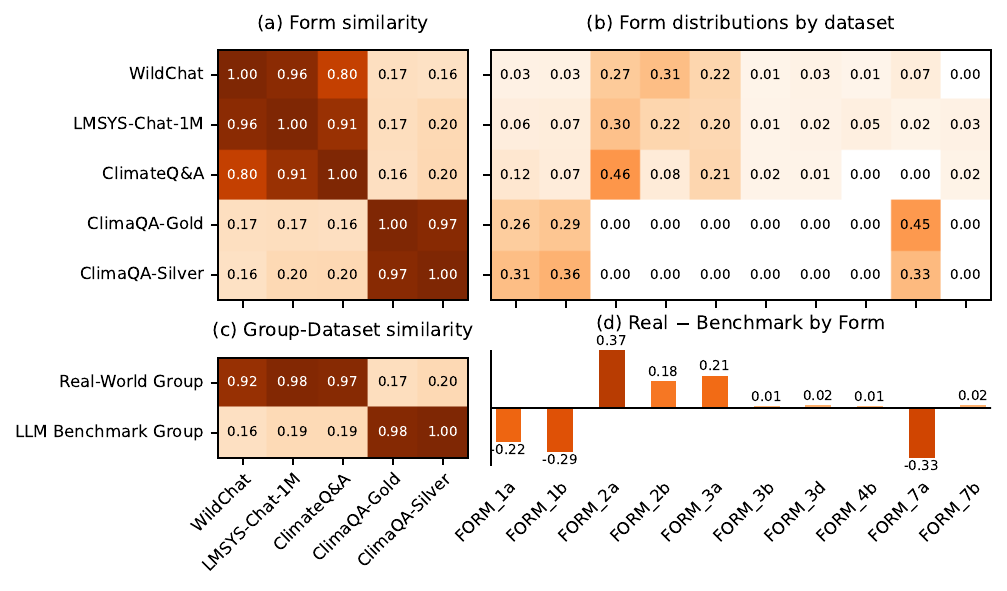}
    \caption{Expected answer forms comparison between Human-to-AI Queries and Human-to-AI Guidance Knowledge.}
    \label{fig:rq1_form}
\end{figure}

For \textbf{Expected Answer Form}, shown in Figure~\ref{fig:rq1_form}, within-group consistency remains high, while between-group differences are substantial. The Benchmark Group focuses on \textit{FORM\_1a. Concise Value(s) / Entity(ies)}, \textit{FORM\_1b. Brief Statement}, and \textit{FORM\_7a. Multiple Choice}. In contrast, the Real-World Group exhibits richer diversity but primarily concentrates on \textit{FORM\_2a. Concise Paragraph}, \textit{FORM\_2b. Detailed Multi-paragraph}, and \textit{FORM\_3a. Item List}, with between-group gaps reaching up to 37\% (Figure~\ref{fig:rq1_form}(d)).

\begin{figure}[t]
    \centering
    \includegraphics[width=\columnwidth,page=1]{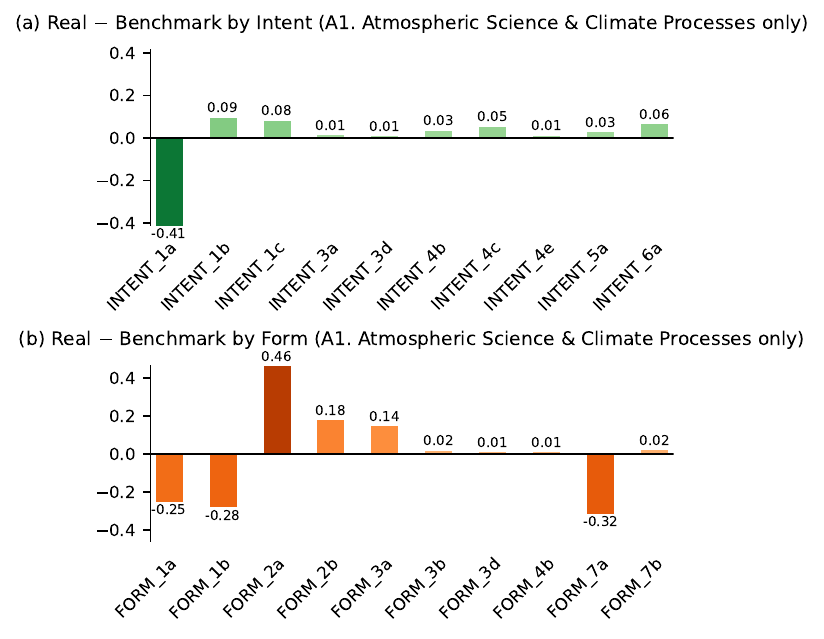}
    \caption{Question-type differences within topic \textit{A1. Atmospheric Science\&Climate Processes}: Real-World queries to LLM vs LLM QA benchmarks. (a) Top-10 intent differences; (b) Top-10 answer form differences.}
    \label{fig:rq1_diff}
\end{figure}

Because the Benchmark Group is highly concentrated in topic \textit{A1}, we further examine whether the observed differences in intent and form are driven solely by topic imbalance. Restricting both groups to topic \textit{A1} (Figure~\ref{fig:rq1_diff}), we find that the differences in intent and form remain robust and align with the patterns observed in the full data (cf.\ Figure~\ref{fig:rq1_intent}(d) and Figure~\ref{fig:rq1_form}(d)).

Taken together, these results show that current LLM climate evaluations exhibit a systematic misalignment with real user needs across the \textit{Topic--Intent--Form} space. Benchmarks concentrate narrowly on climate-science fundamentals, prioritize \textit{Factual} and limited \textit{Conceptual knowledge} via fact-lookup and reasoning-analysis query types (see the intent–knowledge mapping in Figure~\ref{fig:question_type} in Appendix~\ref{app:Taxonomies}), and remain restricted to relatively simple answer forms. 
In contrast, users’ actual knowledge needs span a much broader set of topics (such as policy, transition, and industry action), require higher-level \textit{Procedural} and \textit{Metacognitive knowledge} to support advisory and operational-writing intents, and involve more complex answer forms such as paragraphs and itemized lists.

\begin{figure}[t]
    \centering
    \includegraphics[width=\columnwidth,page=1]{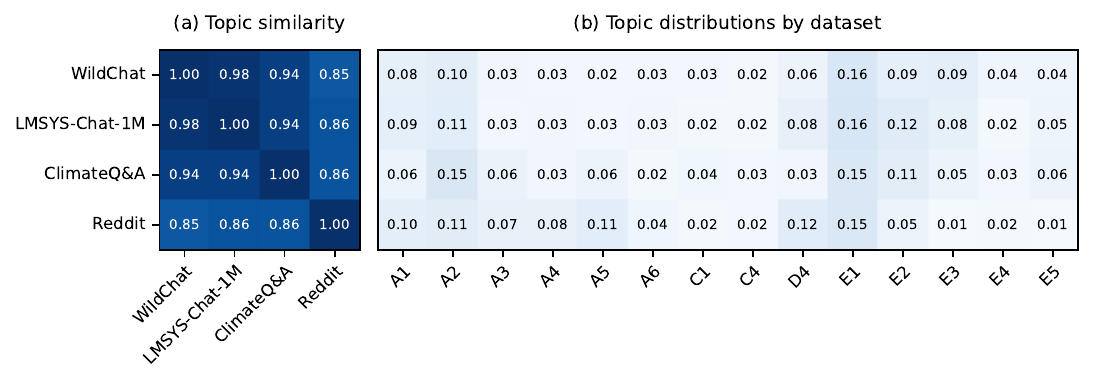}
    \caption{Topic comparison between real-world queries to LLMs and questions to humans. (a) Pairwise topic-distribution similarity across four datasets; (b) Topic distributions for the union of Top-10 topics (by share) across the four datasets. The interpretations and presentations of Figures~\ref{fig:rq2_intent} and~\ref{fig:rq2_form} follow the same logic.}
    \label{fig:rq2_topic}
\end{figure}

\begin{figure}[t]
    \centering
    \includegraphics[width=\columnwidth,page=1]{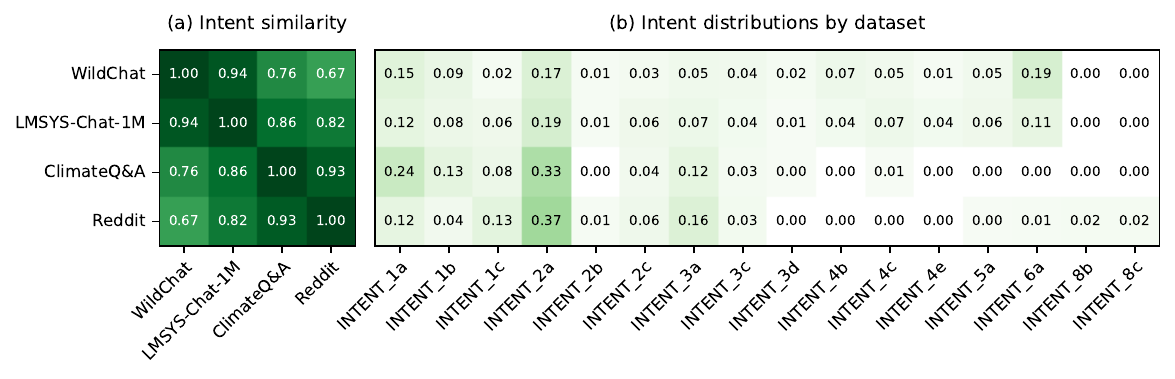}
    \caption{User intent comparison between real-world queries to LLMs and real-world questions to humans.}
    \label{fig:rq2_intent}
\end{figure}

\begin{figure}[t]
    \centering
    \includegraphics[width=\columnwidth,page=1]{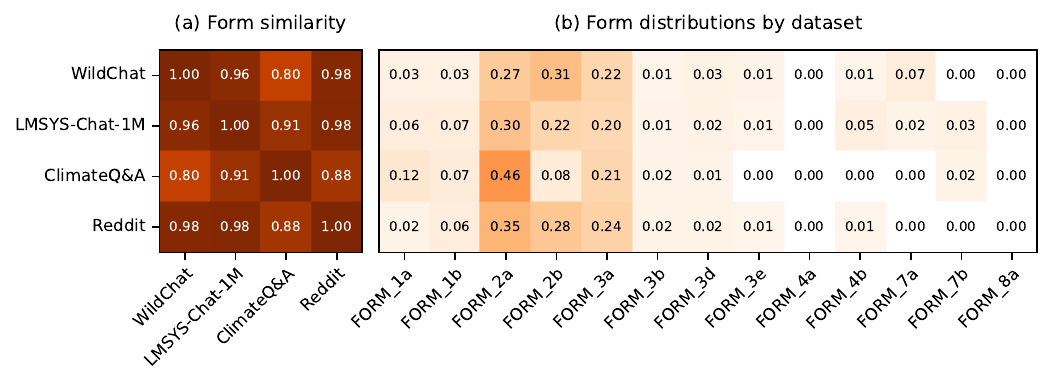}
    \caption{Expected answer form comparison between real-world queries to LLMs and real-world questions to humans.}
    \label{fig:rq2_form}
\end{figure}

\subsection{Similarity in Knowledge Needs}
\label{r:result2}
To verify whether \textbf{human–human knowledge interactions can serve as a reference for climate-related LLM evaluation}, we compare Human-to-AI Queries with Human-to-Human Questions.
Recall that the latter consists of questions posted on Reddit.
At the \textbf{Topic} level, as shown in Figure~\ref{fig:rq2_topic}, the four datasets exhibit high distributional similarity (minimum 85\%), suggesting that users do not systematically avoid or prefer particular topics when seeking answers from different knowledge providers. \textit{E1. Climate Policy, Governance \& Finance Mechanism} and \textit{A2. Greenhouse Gas \& Biogeochemical Cycles} are the most prominent topics (with minimum shares of roughly 15\% and 10\%, peaking at 16\% and 15\%). \textit{A1. Atmospheric Science \& Climate Processes}, \textit{D4. Public Awareness, Communication \& Community Engagement}, and \textit{E2. Energy Transition} also generally maintain nontrivial proportions.

At the \textbf{User Intent} level (Figure~\ref{fig:rq2_intent}), overall patterns are similar, yet differences become more noticeable: \textit{INTENT\_1a. Fact Lookup} is markedly higher in \textit{ClimateQ\&A} (around 24\%), consistent with the intuition that a domain-specialized LLM is frequently used for factual lookup. Meanwhile, \textit{INTENT\_2a. Reasoning / Causal Analysis} is higher in both \textit{ClimateQ\&A} and \textit{Reddit}--the two datasets representing domain-specific climate contexts--about 33\% and 37\%. In contrast, the intent distributions of \textit{WildChat} and \textit{LMSYS-Chat-1M} are more dispersed; relative to the other datasets, these two open-domain conversational datasets contain a higher proportion of \textit{INTENT\_6a. Operational Writing} (i.e., writing/creation tasks).

At the \textbf{Form} level (Figure~\ref{fig:rq2_form}), the four datasets are highly similar, with a general preference for \textit{FORM\_2a. Concise Paragraph}, \textit{FORM\_2b. Detailed Multi-paragraph}, and \textit{FORM\_3a. Item List}.

In summary, users largely focus on the same topics when addressing LLMs versus humans, and consistently prefer well-explained, clearly structured text or itemized forms. However, intent varies by context: domain-specific LLMs and forums emphasize analysis-oriented and fact-based questions (corresponding to \textit{Conceptual} and \textit{Factual knowledge}), while using general-purpose LLMs for climate-related writing remains an important use case (corresponding to \textit{Procedural knowledge}). Recognizing these differences helps tailor evaluation strategies to specific application scenarios.

\subsection{Knowledge Sources Align with Needs}
\label{r:result3}

\begin{figure}[t]
    \centering
    \includegraphics[width=0.62\columnwidth,page=1]{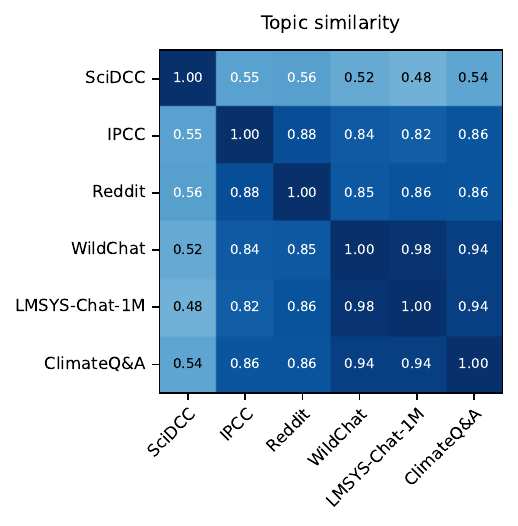}
    \caption{Topic comparison between inform-human knowledge and both real-world questions to human and queries to LLMs.}
    \label{fig:rq3_sim}
\end{figure}

To \textbf{identify high-quality knowledge sources in human–human knowledge interactions that align with real-world knowledge needs}, we analyze the \textbf{Topic} distributions between Human-to-Human Knowledge Provision—represented by \textit{IPCC AR6} and \textit{SciDCC}—and both Human-to-Human Questions and Human-to-AI Queries. Overall, as shown in Figure~\ref{fig:rq3_sim}, 
although \textit{IPCC AR6} is commonly perceived as highly specialized and policy-oriented, and has raised concerns about whether it can meet the public’s knowledge needs, its topic distribution is in fact highly aligned with the human climate knowledge demands (with similarities of 88\%, 84\%, 82\%, and 86\% to Reddit, WildChat, LMSYS-Chat-1M, and ClimateQ\&A, respectively). In contrast, the topic similarity of \textit{SciDCC} appears to be more moderate.

\begin{figure}[t]
    \centering
    \includegraphics[width=\columnwidth,page=1]{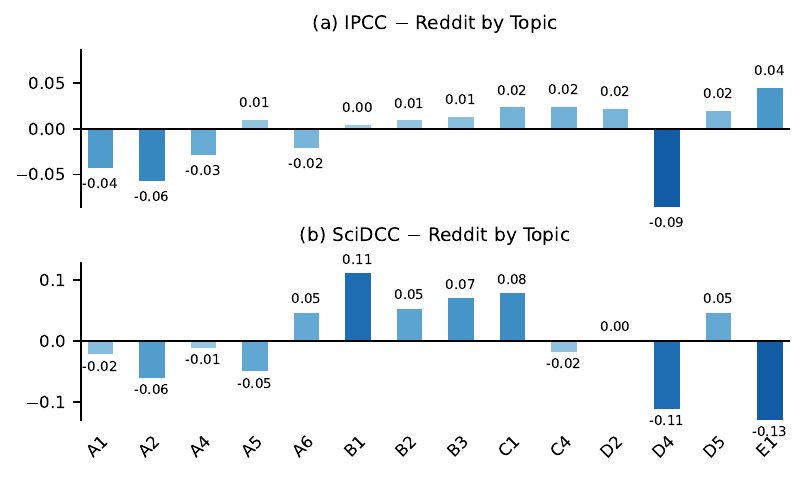}
    \caption{Topic differences: inform-human knowledge vs real-world questions to human. (a) IPCC minus Reddit topic distributions; (b) SciDCC minus Reddit topic distributions. Only shows the union of Top-10 absolute differences.}
    \label{fig:rq3_diff}
\end{figure}

To further leverage these corpora, we compute the distribution differences between \textit{IPCC/SciDCC} and Human needs (Figure~\ref{fig:rq3_diff}). Considering the strong consistency in knowledge demands between Human-to-AI queries and Human-to-Human questions, we simplify the analysis by using Reddit as a proxy for Human needs. We observe the first major contrast: \textit{SciDCC} is more strongly oriented toward \textit{B. Ecological Impacts}, particularly \textit{B1. Biodiversity Loss}. We further identify a shared gap: compared with \textit{Reddit}, both \textit{IPCC} and \textit{SciDCC} devote substantially less attention to \textit{D4. Public Awareness, Communication \& Community Engagement}. This suggests that although these issues are salient to users, they remain underrepresented in knowledge-provision sources.

\subsection{Insights}

Based on the findings in Results~\ref{r:result1},~\ref{r:result2}, and~\ref{r:result3}, we derive the following implications to help design climate-change–related LLM evaluations and improve LLM knowledge provision

\textbf{First, Benchmark Design.}
Benchmark questions should be designed according to the observed distributions of topic, user intent and expected answer form, so as to better reflect real-world user needs. In addition, \textit{IPCC AR6} closely aligns with real-world user topic demands, making it an ideal source for (down-)sampling data when constructing benchmark datasets.

\textbf{Second, Retrieval-Augmented Generation (RAG) Systems.}
Similarly, \textit{IPCC AR6} serves as an excellent retrieval corpus. Its strong alignment with real-user topic distributions and its rich content can reduce redundant data collection and improve retrieval relevance, making it suitable as a minimal knowledge base. This finding also helps explain why some previous studies~\citep{ChatClimate} achieved better performance when using IPCC reports for RAG systems. However, for the topic \textit{D4: Public Awareness, Communication \& Community Engagement}, additional external sources are still required to supplement the retrieval corpus.

\textbf{Finally, Training and Fine-tuning.}
The distributional differences we identified in topics and question types can directly inform the composition of training datasets for climate-specific LLMs. These insights provide guidance for data selection and balancing, thereby improving the alignment between models and real-world knowledge needs.

%% file: Sections/6_Conclusion.tex
\section{Conclusion}

We investigate whether climate change benchmarks for LLMs reflect real-world knowledge needs. By comparing benchmarks with real user--LLM interactions, we find a systematic misalignment across the \textit{Topic--Intent--Form} space: existing benchmarks emphasize science and fact-lookup, whereas real users seek broader, application-oriented knowledge and structured explanations. We further show that human--human and human--AI knowledge-seeking patterns are highly similar, and identify that \textit{IPCC AR6} closely matches real-world topic distributions. This offers guidance for more realistic climate-related LLM evaluation, development and RAG system construction.

%% file: Sections/7_Limitation.tex
\section*{Limitations}

This work focuses on climate change, therefore, its methodology and some of its conclusions may have limited direct generalizability to other domains. In addition, certain data categories are constrained by the availability of high-quality datasets. For example, human--human queries are sourced solely from \textit{Reddit}, primarily because existing public climate-related datasets from pllatform like Twitter (now X) tend to exhibit substantial noise in both content and format. Similarly, the Human-to-AI Guidance Knowledge category relies exclusively on \textit{ClimaQA}, as it is currently one of the few mainstream and relatively well-curated LLM-oriented QA datasets in the climate change domain.

Although we consider the Human-to-AI Queries to be of sufficient scale, and datasets such as \textit{WildChat} and \textit{LMSYS-Chat-1M} cover multilingual and geographically diverse users, with distributions across Topic-Intent-Form showing strong consistency across sources, the data may still exhibit potential biases. In particular, users who consent to sharing their interactions with LLMs may constitute a self-selected population, and English speakers remain dominant in the data. This may result in the underrepresentation or omission of certain populations and their needs.

Furthermore, the construction of the topic taxonomy in this work involves a degree of manual decision-making, which may introduce subjectivity. Although we show in Appendix~\ref{app:2 merge analysis} that directly using the topic list obtained after Topic Merge yields results similar to those after Topic Reassignment in the overall analysis, the taxonomy design itself may still influence the final conclusions.

Finally, the fine-grained topic and question taxonomy increases the complexity of manual annotation and validation, and places higher demands on annotator expertise, which in turn raises the difficulty of thoroughly validating the results.

%% file: Sections/Appendix_1.tex
\section{Taxonomies}
\label{app:Taxonomies}

\begin{figure*}[htbp]
    \centering
    \includegraphics[width=\textwidth, page=1]{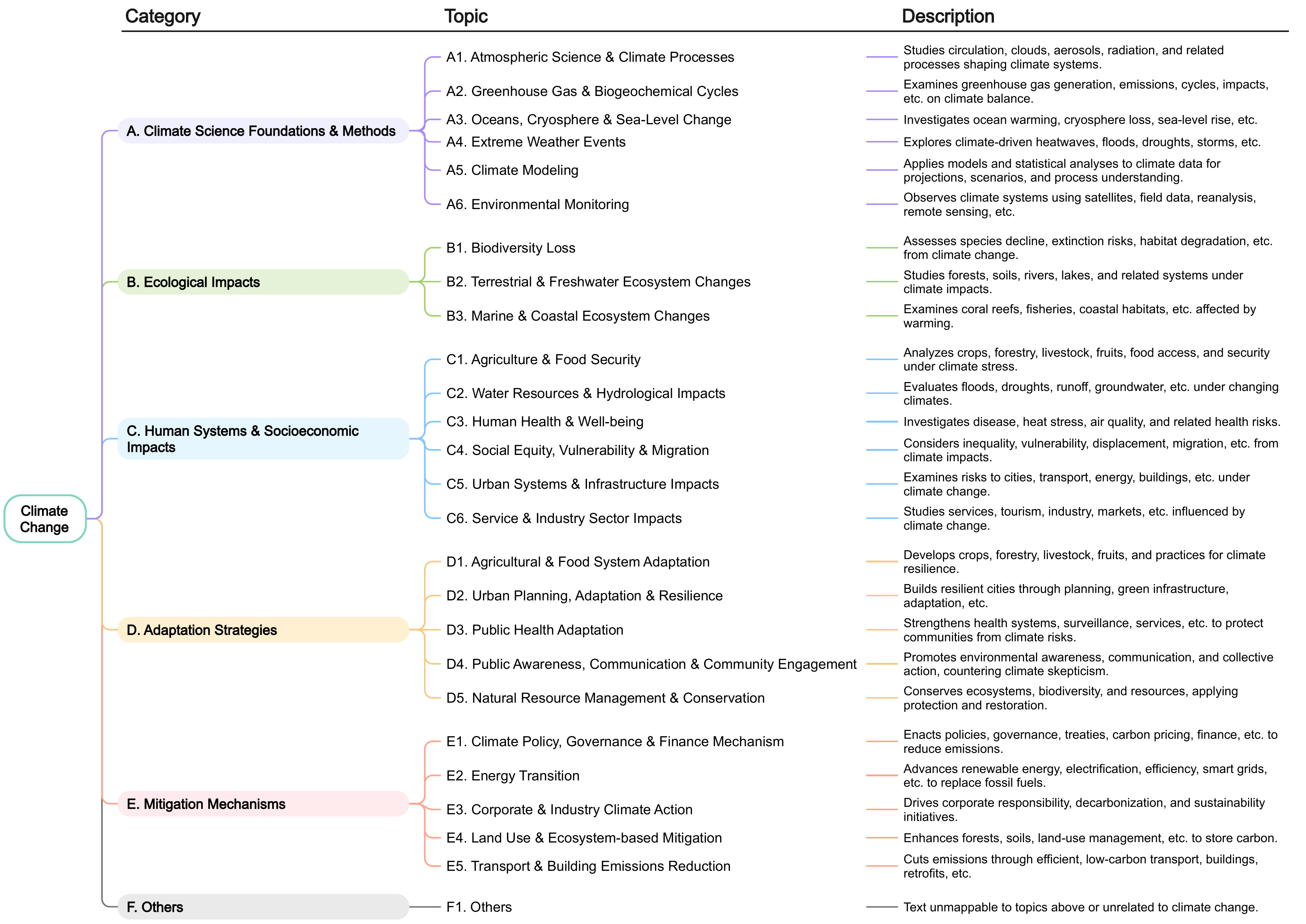}
    \caption{Final Topic Taxonomy of Climate Change.}
    \label{fig:topic_taxonomy}
\end{figure*}

\begin{figure*}[htbp]
    \centering
    \includegraphics[width=\textwidth, page=1]{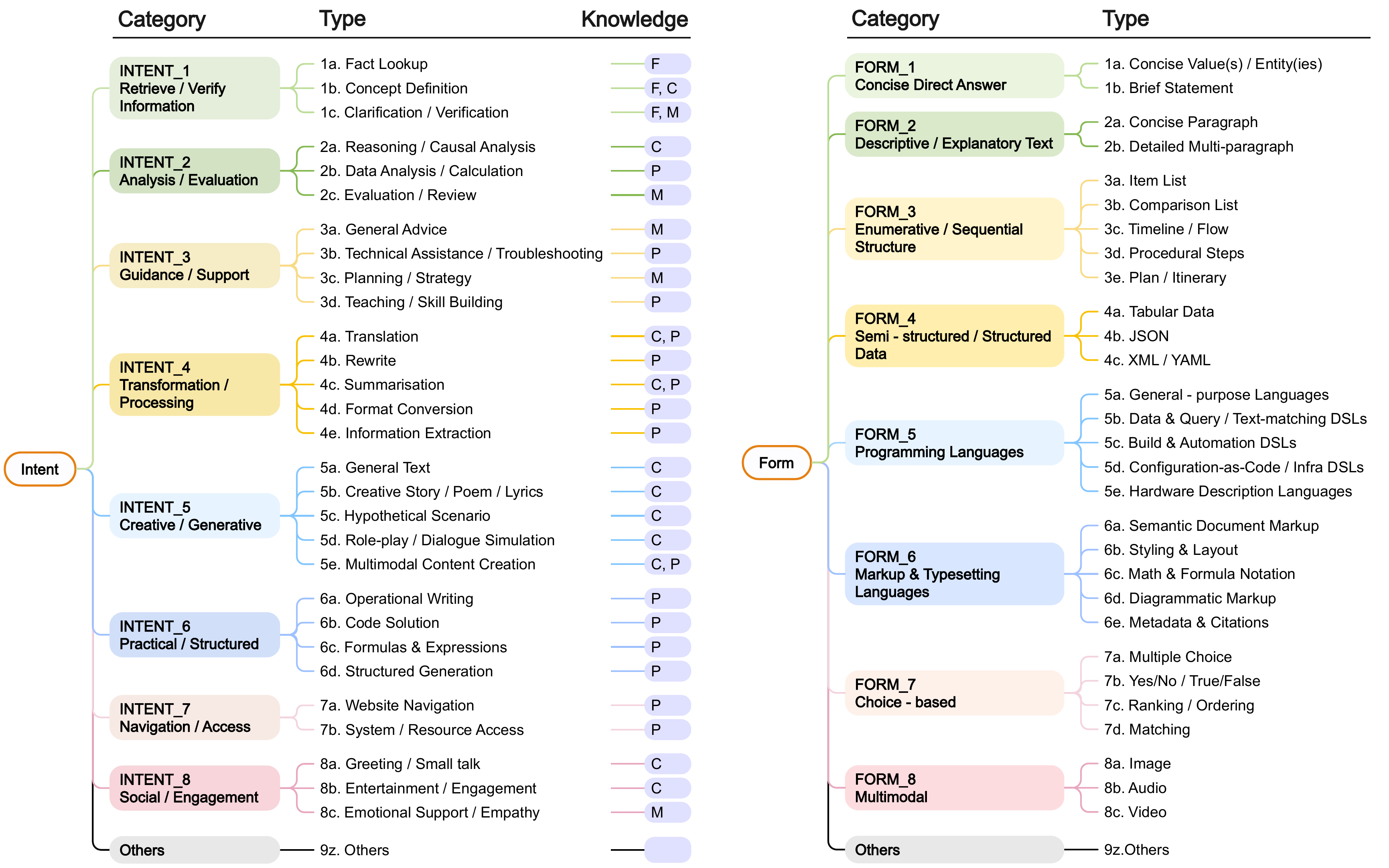}
    \caption{Taxonomy of question types, including user intents the expected answer forms. The Intent Taxonomy also contains a “Knowledge” label indicating the type of knowledge required by an LLM for each intent: \textbf{F} for Factual, \textbf{C} for Conceptual, \textbf{P} for Procedural, and \textbf{M} for Metacognitive. Each category also has an \textit{z. Others} type, which is omitted from the figure.}
    \label{fig:question_type}
\end{figure*}

Figures~\ref{fig:topic_taxonomy} and~\ref{fig:question_type} present the final taxonomies developed and used for data annotation in our work. Specifically, Figure~\ref{fig:topic_taxonomy} illustrates the climate change \textit{topic taxonomy}, while Figure~\ref{fig:question_type} shows the \textit{question type taxonomy}.

The topic taxonomy is organized as a two-level hierarchical structure. It contains five primary categories—\textit{A. Climate Science Foundations \& Method}, \textit{B. Ecological Impacts}, \textit{C. Human Systems \& Socioeconomic Impacts}, \textit{D. Adaptation Strategies}, and \textit{E. Mitigation Mechanisms}—comprising a total of 25 fine-grained subtopics. In addition, we introduce an auxiliary category \textit{F. Others} to capture a portion of climate-related data not covered by the main taxonomy, as well as irrelevant samples. Data labeled as \textit{Others} are excluded from the final analysis.

The question type taxonomy consists of two complementary two-level classification schemes: one describing \textit{user intent} and the other specifying the \textit{expected answer form}. Each scheme includes eight core categories and 29 fine-grained subtypes. Additionally, both taxonomies contain a global \textit{Others} category and an extra \textit{others} subtype within each core category to account for edge cases.

Furthermore, we incorporate the knowledge dimension framework from Bloom’s taxonomy~\citep{Knowledge_Taxonomy}, including \textit{Factual knowledge} (basic facts and terminology), \textit{Conceptual knowledge} (relationships among concepts, principles, and theories), \textit{Procedural knowledge} (methods and processes for performing tasks), and \textit{Metacognitive knowledge} (awareness and regulation of one’s own cognition, including strategies for learning). Since the taxonomy was originally developed to describe human learning processes, we adapt and operationalize these categories to better reflect knowledge usage in LLMs. Specifically, we interpret \textit{Factual knowledge} as the recall of basic facts and definitions, \textit{Conceptual knowledge} as the ability to capture relationships among concepts (e.g., causal or relational reasoning), \textit{Procedural knowledge} as knowledge of how to execute tasks or processes, and \textit{Metacognitive knowledge} as the capacity for planning, strategy selection, and decision-making during problem solving. This extension is intended to provide an operational categorization of knowledge needed by users in LLM outputs rather than to model human cognition directly.

These labels indicate the primary and most representative knowledge requirement for addressing each intent type. For example, \textit{1a. Fact Lookup} primarily rely on \textit{Factual} knowledge (F), whereas \textit{2a. Reasoning / Causal Analysis} typically depends on \textit{Conceptual} knowledge (C). It is important to note that we annotate only the most distinctive and discriminative knowledge dimension for each category. Additional supporting knowledge requirements and the detailed mapping rationale are documented in Appendix~\ref{app: knowledge_taxonomy_reason}.

%% file: Sections/Appendix_2.tex
\section{Method and Data Details}
\label{app:repro}

\subsection{More details of the dataset}
\label{app:auxillary}
\paragraph{Auxiliary Corpora} The \textbf{Auxiliary Corpora} included three datasets: \textit{Climate-FEVER}~\citep{Climate_FEVER}, a fact verification dataset; \textit{Environmental Claims}~\citep{Environmental_Claims}, which focuses on corporate climate claims; and \textit{ClimSight}~\citep{ClimSight}, a climate question–answering dataset. More details in Table~\ref{tab:auxillary}.

\paragraph{Data Consent}

We use a combination of publicly available datasets and data collected via official APIs, and adhere to their respective licenses and terms of use. All datasets used in this work are obtained from publicly released sources. We rely on the original data collection procedures and consent processes as described by their authors (in accordance with their respective licenses, e.g., Apache-2.0 License, BSD 3-Clause License,), as specified in their official releases.), and use these datasets strictly for research purposes. Reddit data is collected using the official Reddit API in compliance with the platform's terms of service. The data consists only of publicly available posts. We do not collect private or deleted content. The IPCC AR6 reports (Working Groups I, II, and III) are publicly available documents intended for open dissemination and are used for research purposes only.

\paragraph{Sensitive or Offensive Content} The study uses publicly available datasets (e.g., \textit{WildChat}, \textit{LMSYS-Chat}, \textit{ClimateQ\&A}). These datasets consist of user queries and may contain naturally occurring personally identifiable information or offensive content. We rely on the original dataset providers' collection and filtering procedures and do not introduce additional sensitive data. This work does not involve direct interaction with human subjects, and no attempt is made to identify individuals represented in the data.

\begin{table*}[htbp]
\centering
\begin{adjustbox}{max width=\textwidth}
\scriptsize
\setlength{\tabcolsep}{4pt}
\renewcommand{\arraystretch}{1.15}
\begingroup
\setlength{\parindent}{0pt}
\setlength{\aboverulesep}{0.6ex}
\setlength{\belowrulesep}{0.6ex}
\setlength{\cmidrulesep}{0.4ex}
\begin{tabular}{
  >{\RaggedRight\arraybackslash}p{0.15\linewidth}  % Category
  >{\RaggedRight\arraybackslash}p{0.17\linewidth}  % Dataset
  >{\RaggedRight\arraybackslash}p{0.20\linewidth}  % Task/Format
  >{\RaggedRight\arraybackslash}p{0.41\linewidth}  % Brief Description
  >{\RaggedRight\arraybackslash}p{0.07\linewidth}   % Count
}
\toprule
\textbf{Category} & \textbf{Dataset} & \textbf{Task/Format} & \textbf{Brief Description} & \textbf{Count} \\
\midrule

% Category 5
\multirow{3}{*}{\makecell[l]{\textit{Auxiliary Corpora}}} &
Climate-FEVER &
Fact Verification &
Internet climate-related claims for factuality assessment. & 1,452 \\
\cmidrule[0.6pt](lr){2-5}
& Environmental Claims &
Claim Detection \brk{Corporate} &
Company environmental statements from annual/sustainability reports; corporate perspective. & 1,270 \\
\cmidrule[0.6pt](lr){2-5}
& ClimSight &
QA Dataset &
A question-answering dataset related to climate change. & 2,998 \\
\bottomrule
\end{tabular}
\endgroup
\end{adjustbox}

\caption{An overview of the 3 Auxiliary Corpora used in this study.}
% in Data Collection and Topic Identification stages.}
\label{tab:auxillary}
\end{table*}

\newtcolorbox{FullWidthBox}[1][]{
  breakable,
  enhanced,
  colback=gray!5,
  colframe=black,
  width=\textwidth,
  left=6pt,right=6pt,top=6pt,bottom=6pt,
  #1
}

\tcbset{
  mybox/.style={
    colback=orange!5,
    colframe=orange!70!black,
    breakable,
    boxrule=0.5pt,
    arc=2pt,
    outer arc=2pt
  }
}
\lstset{
  basicstyle=\ttfamily\small,
  breaklines=true,
  breakatwhitespace=false,
  columns=fullflexible,
  keepspaces=true,
}

\subsection{LLM Prompts}
The prompt below omits the example and taxonomy. In the following prompt, \{subject\}, \{n\}, and \{m\} correspond to \textit{Climate Change}, 4, and 20, respectively.
% The full version is available in the code repository.

\label{app:Prompt 1}
\begin{tcolorbox}[mybox, title={1. Topic Identification: Initial Topic Generation}]
\begin{lstlisting}
You are an expert research curator specialising in "{subject}". Your task is to identify the main topics touched on by the text below, and for each topic provide a brief (<= {m} words) explanation.
The texts you need to identify will come from various sources, including reports, real-world user queries, online Q&A, AI-generated questions, social media posts, and more. At the same time, these texts may appear in any language (including English, Chinese, Russian, French, etc.). You need to detect the language, but always produce the output in English.

GUIDELINES:

1. Think of a "topic" as a high-level theme-but neither too abstract (e.g. "science") nor too fine-grained (e.g. "NO2 emissions from Euro-6 diesel cars in Canberra on 12 Jan 2021"). Some examples of granularity guide fot generated topics: Climate Change: Extreme Weather; Climate Change: Climate Modeling; Climate Change: Urban Planning; Climate Change: Economic Impacts.
2. Express each topic in the form "{subject}: <related_domain>" where `<related_domain>` is:
- A short, self-contained descriptor (no more than n={n} words), such as "entrepreneurship", "agriculture", "politics", or "natural ecology";
- A concrete subdomain within {subject}, or an interdisciplinary area related to {subject}.
3. Return between 1 and 3 topics. If the content is no plausible connection to {subject}, return exactly:[  {"topic":"Irrelevant Data","explanation":"None" }], not"Climate Change: Irrelevant Data".
4. Output a pure array of objects in English, each with:
-"topic": a string like"Climate Change: Economic Phenomena";
-"explanation": a string that explains the meaning of the topic in the context of {subject}, using no more than m = {m} words.

[Examples (4 positive and 2 negative).]
\end{lstlisting}
\end{tcolorbox}

% ----------------- Step 2 -----------------
\label{app:Prompt 2}
\begin{tcolorbox}[mybox, title={2. Topic Identification: Iterative Topic Merging}]
\begin{lstlisting}
You are a meticulous topic curator. Each topic is represented as a JSON object with two keys:
-"topic": the topic label
-"explanation": a short description of what that topic covers

The system will pass you a JSON array where the first element is the parent topic, and the remaining elements are the candidate topics to consider. **Each object also includes an"id" field which is the 1-based row index (as a string). Use these"id" values for selection:**
[
  {"id":"1","topic":"Climate Change: ParentLabel","explanation":"Parent explanation here" },
  {"id":"2","topic":"Candidate A","explanation":"Explanation A" },
  {"id":"3","topic":"Candidate B","explanation":"Explanation B" },
  ...
]

Your job is to merge those candidates whose both topic and explanation clearly match the parent concept.

Return a JSON object with exactly three keys:

{
 "merged_ids": [ /* array of candidate IDs (strings like"2","3") to merge */ ],
 "parent_topic":     "...",    /* either unchanged or a clearer single topic label */,
 "parent_explanation":"..."    /* a concise explanation for the chosen parent_topic */
}

Rules:
1. ONLY merge when the candidate's topic label is a near-synonym of the parent's label (same concept, same granularity) and the candidate's explanation bidirectionally entails the parent's explanation (each would still be true if swapped). Do not merge if the candidate is a sibling area.
2. If unsure, do not merge! Just returning an empty"merged_ids" over a risky merge; only merge high-confidence near-duplicates.
3. Maintain the same semantic level (e.g., do not merge into a broader or narrower topic).
4. Do not include the parent itself in"merged_ids".
5. You must only select items from the provided list and return their"id" values in"merged_ids" (do not invent new IDs).
6. If no candidates should merge, return an empty array for"merged_ids" and repeat the original"parent_topic" and"parent_explanation".
7. You may revise the"parent_topic" to improve clarity or specificity, but it must follow the required topic naming format below.

Topic Naming Format (Mandatory for parent_topic and all candidates):
1. Each topic must be expressed in the form:"{subject}: <related_domain>" where <related_domain> must be:
- A short, self-contained descriptor (no more than n={n} words), such as "entrepreneurship", "agriculture", "politics", or "natural ecology";
- A concrete subdomain within {subject}, or an interdisciplinary area related to {subject}.
- Avoid vague or overly general terms like "issues", "overview", or "aspects".
2. Each topic's explanation explains the general meaning of the topic in the context of {subject}, using no more than m = {m} words.

[A complete example of topic merging.]
\end{lstlisting}
\end{tcolorbox}

% ----------------- Step 3 -----------------
\label{app:Prompt 3}
\begin{tcolorbox}[mybox, title={3. Topic Identification: Topic Reassignment}]
\begin{lstlisting}
You are an expert in climate change science.
Classify the input text into at most 3 core topics based on the taxonomy below.

**Decision Rules:**
1. Identify only **core, directly relevant** topics (do not assign marginal or weakly related ones).
2. Return up to 3 topics; fewer is acceptable.
3. If the text **does not clearly map to any topic or unrelated to climate change**, return Others only.
   - When classifying as Others, return only one item:
   [
     {"topic":"F1. Others" }
   ]
4. Do not create new topics. Use only the taxonomy provided.

Output Format (STRICT)
Return ONLY a JSON array of 1-3 objects, each with a single field:
[
  {"topic":"A5. Climate Modeling" },
  {"topic":"E1. Climate Policy, Governance & Finance Mechanism" }
]

[The complete, two-level topic taxonomy (including explanations for each topic).]
[Examples (3 positive and 1 negative).]
\end{lstlisting}
\end{tcolorbox}

% ----------------- Step 4 -----------------
\label{app:Prompt 4}

\begin{tcolorbox}[mybox, title={4. Question Type Classification}]
\begin{lstlisting}
You are a query classification expert.
Given a user query Q, classify it along two independent axes:
AXIS A - INTENT (What the user wants to achieve)
AXIS B - FORM (The most appropriate form of the answer)
The texts you need to identify will come from various sources, including real-world user queries, online Q&A, AI-generated questions, social media posts, and more. At the same time, these texts may appear in any language (including English, Chinese, Russian, French, etc.). You need to detect the language, but always produce the output in English.

===============================
Classification rules
===============================

1. For both INTENT and FORM, assign only one main category if possible.
2. If the query reasonably fits multiple categories (e.g., user expects both evaluation and advice), assign all applicable categories but no more than 3.
3. When ambiguous in FORM (e.g., concept explanation may be concise or detailed), assign multiple categories as needed but no more than 3.
4. If multiple categories are assigned for either INTENT or FORM, list them together in an array in the output, in priority order (most relevant first).
5. If no category fits, assign the appropriate Others category (marked with "z").
6. Output must in English and format must be:
   {
     "intent": ["INTENT_xx. xxx", "INTENT_yy. yyy", ...],
     "form": ["FORM_xx. xxx ", "FORM_yy. yyy", ...]
   }
7. Do not output any text outside the JSON structure.

[The complete, two-level intent and form taxonomy (including brief explanations for each intent/form and linguistic clues (e.g., common query patterns) for each intent).]
[Examples (4 positive).]
\end{lstlisting}
\end{tcolorbox}

\subsection{LLM usage and configurations.}
\label{app:LLMuse}
For \textbf{Filtering Data} from \textit{WildChat} and \textit{LMSYS-Chat-1M}, we used the \texttt{Qwen-3 30B} model~\citep{Qwen3} (temperature=0, reasoning mode disabled).
For \textbf{Initial Topic Generation}, we employed \texttt{GPT-4.1-mini}~\citep{GPT-4.1-mini} (temperature=0.2).
For \textbf{Iterative Topic Merging}, we used \texttt{GPT-4.1-mini} (temperature=0.2) or \texttt{GPT-5-mini} (minimal reasoning effort) depends on merge settings. Both \textbf{Topic Reassignment} and \textbf{Question Type Classification} were performed using \texttt{GPT-5-mini}~\citep{GPT-5-mini} (low reasoning effort).

\subsection{Additional Details on Topic Modeling.}
\label{app:add_details_topic}
\textbf{Initial Topic Generation.}
To ensure consistency in the generated topics, we impose a constrained naming format and granularity requirement within the prompt. Each topic follows the structure 
\textit{\{subject\}: <related-domain>}, 
where \textit{<related-domain>} denotes a specific area closely associated with the research subject (in our case, climate change). Empirically, this naming convention helps align the generated topics more closely with the semantic scope of the target subject. 

To further control topic granularity, we impose two additional constraints: the \textit{related-domain} component is limited to at most four words, and the accompanying explanation is restricted to at most twenty words. These restrictions prevent the model from generating overly fine-grained topics that capture idiosyncratic details of individual samples rather than generalizable thematic concepts.

\textbf{Iterative Topic Merging.}
In the iterative merging stage, we first sort the initial topic list by frequency in descending order. In each iteration, the most frequent topic that has not yet been merged is selected as the anchor topic. This strategy is motivated by the intuition that high-frequency topics tend to be more representative of the corpus and therefore serve as effective anchors for consolidating semantically related topics.

For each anchor topic, we retrieve the top-$k$ most similar candidate topics as potential merge targets, where $k= \min\!(10,\max(1,\lfloor \textit{Topic List Size}/10 \rfloor))$. Empirically, this value provides a practical balance between coverage and quality: a smaller $k$ reduces the efficiency of the merging process, while a larger $k$ tends to degrade LLM performance during the semantic comparison stage.

Although embedding similarity provides a useful signal for identifying potential candidates, it may still be affected by biases or coverage limitations in the embedding model’s training data. As a result, some topics with similar vector representations may correspond to conceptually unrelated themes. For example, topics such as ``Drought Risk'' and ``Urban Planning'' may appear close in the embedding space despite describing different concepts.

After each round of merging, the remaining topics are carried forward to the next iteration. The process continues until one of the following stopping conditions is satisfied: 
(1) \textbf{Inactivity}, where no merges occur during the current iteration, indicating that the process has converged; or 
(2) \textbf{Spread-based criteria}, where the semantic similarity among the remaining topics is sufficiently low (mean similarity $<0.3$ or maximum similarity $<0.5$), suggesting that the remaining topics are already sufficiently diverse and further merging is unlikely to be beneficial. 

In practice, the merging process typically terminates due to the inactivity condition. The topics produced in the \textit{Iterative Topic Merging} stage retain the same naming format as those generated in the \textit{Initial Topic Generation} stage. The pseudo-code of the topic merging algorithm is presented in Algorithm~\ref{alg:iterative_topic_merging}, and the experiment analyzing the topic list obtained after topic merging is presented in Appendix~\ref{app:2 merge analysis}.

\begin{algorithm*}[htbp]
\caption{Iterative Topic Merging}
\label{alg:iterative_topic_merging}
\KwIn{
Initial topic set
\(\mathcal{T}=\{(t_i,e_i,c_i,\mathbf{v}_i)\}_{i=1}^{N}\),
where \(t_i\) is the topic label, \(e_i\) its explanation,
\(c_i\) its frequency, and \(\mathbf{v}_i\) its embedding;
maximum batch size \(B\);
stopping thresholds \(\theta_{\mathrm{mean}}\) and \(\theta_{\mathrm{max}}\).
}
\KwOut{
A refined topic set \(\mathcal{T}^{*}\) and a hierarchical merge tree \(\mathcal{M}\).
}

\BlankLine
Collapse topics with identical normalized \((t_i,e_i)\) by summing counts and recomputing embeddings\;

\Repeat{
    no merge occurs in the current round
    \textbf{or}
    the mean inter-topic similarity is below $\theta_{\text{mean}}$
    \textbf{or}
    the maximum inter-topic similarity is below $\theta_{\text{max}}$
}{
    Sort \(\mathcal{T}\) in descending order of \(c_i\)\;
    Mark all topics as unmerged\;
    Initialize \(\mathcal{T}_{\mathrm{new}} \leftarrow \varnothing\) and \(\mathcal{M}_{\mathrm{round}} \leftarrow \varnothing\)\;
    Set \(\texttt{any\_merge} \leftarrow \texttt{false}\)\;

    \ForEach{unmerged topic \(p \in \mathcal{T}\) in sorted order}{
        \If{\(p\) is locked}{
            Move \(p\) to \(\mathcal{T}_{\mathrm{new}}\) and record a self-link in \(\mathcal{M}_{\mathrm{round}}\)\;
            mark \(p\) as merged and \Continue\;
        }

        Let \(\mathcal{U}\) be the set of unmerged, unlocked topics excluding \(p\)\;
        \If{\(\mathcal{U}=\varnothing\)}{
            Move \(p\) to \(\mathcal{T}_{\mathrm{new}}\) and record a self-link in \(\mathcal{M}_{\mathrm{round}}\)\;
            mark \(p\) as merged and \Continue\;
        }

        Compute cosine similarities between \(p\) and all topics in \(\mathcal{U}\)\;
        Select the top-\(k\) most similar candidates, where
        \[
        k=\min\!\bigl(B,\max(1,\lfloor |\mathcal{T}|/10 \rfloor)\bigr).
        \]

        Query the LLM with \(p\) and the selected candidates\;
        The LLM returns a subset \(\mathcal{C}\) of candidates to merge with \(p\),
        together with an updated parent topic label and explanation \((t^\star,e^\star)\)\;

        Let \(\mathcal{S}=\{p\}\cup \mathcal{C}\)\;
        \If{\(|\mathcal{S}|>1\)}{
            \(\texttt{any\_merge} \leftarrow \texttt{true}\)\;
        }

        Construct a new parent topic from \(\mathcal{S}\):
        \[
        c^\star=\sum_{x\in\mathcal{S}} c_x,\qquad
        \mathbf{v}^\star=
        \mathrm{Normalize}\!\left(
        \frac{\sum_{x\in\mathcal{S}} c_x\mathbf{v}_x}{\sum_{x\in\mathcal{S}} c_x}
        \right).
        \]
        Append \((t^\star,e^\star,c^\star,\mathbf{v}^\star)\) to \(\mathcal{T}_{\mathrm{new}}\)\;
        Record the parent--children relation in \(\mathcal{M}_{\mathrm{round}}\)\;
        Mark all topics in \(\mathcal{S}\) as merged\;
    }

    Add self-links for any uncovered topics, if needed\;
    Deduplicate \(\mathcal{T}_{\mathrm{new}}\) by identical normalized \((t,e)\), summing counts and recomputing embeddings\;
    Update the global merge tree \(\mathcal{M}\) using the deduplicated parent topics\;
    Set \(\mathcal{T}\leftarrow \mathcal{T}_{\mathrm{new}}\)\;

    Compute the mean and maximum pairwise cosine similarities among topics in \(\mathcal{T}\)\;
}
\Return{\(\mathcal{T},\mathcal{M}\)}\;
\end{algorithm*}

\paragraph{Topic Reassignment.}
In the Topic Reassignment stage, we use the data labeled under the S5 configuration of the topic merging experiments (see Appendix~\ref{app:topic_merge_settings}) as the basis for relabeling. From this dataset, we remove 4,645 samples that were labeled solely as \textit{``Irrelevant Data''} in order to avoid unnecessary computational cost during reassignment. The S5 configuration is selected because, based on a qualitative inspection of the topic merging trajectories, it exhibits the most coherent and well-structured merging behavior among the six experimental settings (see the visualization in Appendix~\ref{app:topic_merging_tree}). 

After filtering out the irrelevant samples, a total of 46,346 samples are reassigned to topics. Among these, 42,261 samples correspond to core-topic categories after excluding 4,085 samples labeled as \textit{``F. Others''}.

\subsection{Experimental Configurations for Six Topic Merging Runs}
\label{app:topic_merge_settings}
Table~\ref{tab:topic_merge_settings} summarizes the six experimental configurations of topic merging pipeline. Each configuration differs in (i) the textual input selected for embedding extraction (topic only vs. topic name + explanation), (ii) the embedding model (all-MiniLM-L6-v2 or Qwen3-Embedding-4B) employed~\citep{all-MiniLM-L6-v2,Qwen3Embedding}, and (iii) the LLM  for executing the topic merging step. The last column reports the final number of merged topics obtained from the initial pool of 10,730 topics.

\begin{table*}[h]
\centering
\begin{adjustbox}{max width=\textwidth}
\scriptsize
\setlength{\tabcolsep}{4pt}
\renewcommand{\arraystretch}{1.5}
\begin{tabular}{
  >{\RaggedRight\arraybackslash}p{0.12\linewidth}  % Setting
  >{\RaggedRight\arraybackslash}p{0.26\linewidth}  % Text for embedding
  >{\RaggedRight\arraybackslash}p{0.21\linewidth}  % Embedding model
  >{\RaggedRight\arraybackslash}p{0.18\linewidth}  % Merging model
  >{\RaggedRight\arraybackslash}p{0.13\linewidth}  % #Merged topics
}
\hline
\textbf{Setting ID} & \textbf{Text for embedding} & \textbf{Embedding model} & \textbf{Merging model} & \textbf{\# Merged topics} \\
\hline
S1 & Topic only & all-MiniLM-L6-v2 & gpt-4.1-mini & 722 \\

S2 & Topic only & all-MiniLM-L6-v2 & gpt-5-mini   & 286 \\

S3 & Topic name + explanation & all-MiniLM-L6-v2 & gpt-4.1-mini & 540 \\

S4 & Topic name + explanation & all-MiniLM-L6-v2 & gpt-5-mini   & 187 \\

S5 & Topic name + explanation & Qwen3-Embedding-4B & gpt-4.1-mini & 411 \\

S6 & Topic name + explanation & Qwen3-Embedding-4B & gpt-5-mini   & 170 \\
\hline
\end{tabular}
\end{adjustbox}
\caption{Six experimental configurations of the topic merging pipeline.}
\label{tab:topic_merge_settings}
\end{table*}

\subsection{Manual Topic Reassignment Details}
\label{app:manual_topic_reassignment_details}
To construct a comprehensive taxonomy, we collected topic candidates from each of the six experimental configurations. The selected lists are shown below. Table~\ref{tab:topic_mapping} maps the selected topics from six experimental configurations (S1-S6) to the final taxonomy, grouped by higher-level categories for clarity. 
\begin{tcolorbox}[colback=gray!5,colframe=black!40,
                  boxrule=0.3pt,arc=2pt,
                  left=4pt,right=4pt,top=4pt,bottom=4pt,
                  breakable,
                  title=\textbf{Setting~1. Amount=23}]
Climate Change: Agriculture, Climate Change: Air Pollution, Climate Change: Atmospheric Science, Climate Change: Biodiversity Conservation, Climate Change: Climate Modeling, Climate Change: Corporate Sustainability, Climate Change: Cryosphere Dynamics, Climate Change: Environmental Impact, Climate Change: Extreme Weather Events, Climate Change: Health Impacts, Climate Change: Hydrological Change, Climate Change: Mitigation Strategies, Climate Change: Policy \& Governance, Climate Change: Public Perception, Climate Change: Regional Climate Impacts, Climate Change: Renewable Energy, Climate Change: Economic Impacts, Climate Change: Greenhouse Gas Emissions, Climate Change: Social Vulnerability, Climate Change: Societal Transformation, Climate Change: Adaptation Strategies, Climate Change: Energy Transition, Climate Change: Biogeochemical Cycles.
\end{tcolorbox}

\begin{tcolorbox}[colback=gray!5,colframe=black!40,
                  boxrule=0.3pt,arc=2pt,
                  left=4pt,right=4pt,top=4pt,bottom=4pt,
                  breakable,
                  title=\textbf{Setting~2. Amount=22}]
Climate Change: Agriculture, Climate Change: Atmospheric Science, Climate Change: Biological Responses, Climate Change: Climate Modeling, Climate Change: Climate Variability, Climate Change: Cloud Processes, Climate Change: Community Engagement, Climate Change: Energy Systems, Climate Change: Extreme Weather, Climate Change: Fossil Fuels, Climate Change: Geophysical Events, Climate Change: Governance, Climate Change: Inventory Methodologies, Climate Change: Mitigation Strategies, Climate Change: Public Health, Climate Change: Radiative Forcing, Climate Change: Regional Policy, Climate Change: Economic Assessment, Climate Change: Economic Structures, Climate Change: Tourism Impacts, Climate Change: Transportation, Climate Change: Urban Planning.
\end{tcolorbox}

\begin{tcolorbox}[colback=gray!5,colframe=black!40,
                  boxrule=0.3pt,arc=2pt,
                  left=4pt,right=4pt,top=4pt,bottom=4pt,
                  breakable,
                  title=\textbf{Setting~3. Amount=25}]
Climate Change: Agricultural Biotechnology, Climate Change: Agriculture, Climate Change: Atmospheric Processes, Climate Change: Biodiversity Loss, Climate Change: Climate Modeling, Climate Change: Corporate Sustainability, Climate Change: Cryosphere Changes, Climate Change: ENSO Dynamics, Climate Change: Energy Transition, Climate Change: Environmental Monitoring, Climate Change: Extreme Weather, Climate Change: Oceanography, Climate Change: Policy \& Governance, Climate Change: Public Awareness, Climate Change: Temperature Trends, Climate Change: Tourism Impacts, Climate Change: Water Resources, Climate Change: Adaptation Strategies, Climate Change: Climate Finance, Climate Change: Community Engagement, Climate Change: Evolutionary Biology, Climate Change: Financial Risk, Climate Change: Human Health and Well-being, Climate Change: Sustainable Development, Climate Change: Urban Planning.
\end{tcolorbox}

\begin{tcolorbox}[colback=gray!5,colframe=black!40,
                  boxrule=0.3pt,arc=2pt,
                  left=4pt,right=4pt,top=4pt,bottom=4pt,
                  breakable,
                  title=\textbf{Setting~4. Amount=31}]
Climate Change: Agriculture, Climate Change: Biodiversity Conservation, Climate Change: Carbon Cycle, Climate Change: Climate Modeling, Climate Change: Climate Science, Climate Change: Cryosphere Changes, Climate Change: Emissions Mitigation, Climate Change: Expert Judgment, Climate Change: Extreme Weather, Climate Change: Fossil Fuel Extraction, Climate Change: Hydrological Impacts, Climate Change: Industry Sustainability, Climate Change: Marine Ecology, Climate Change: Policy \& Governance, Climate Change: Sectoral Impacts, Climate Change: Social Equity, Climate Change: Data \& Observation, Climate Change: Human-Environment Interaction, Climate Change: Natural Disturbances, Climate Change: Public Health, Climate Change: Electrification Impacts, Climate Change: Sustainable Development, Climate Change: Transportation Emissions, Climate Change: Indoor Air Quality, Climate Change: Radiative Forcing, Climate Change: Socioeconomic Policy, Climate Change: Urban Heat Mitigation, Climate Change: ESG Investing, Climate Change: Environmental Pollution, Climate Change: Net Zero Targets.
\end{tcolorbox}

\begin{tcolorbox}[colback=gray!5,colframe=black!40,
                  boxrule=0.3pt,arc=2pt,
                  left=4pt,right=4pt,top=4pt,bottom=4pt,
                  breakable,
                  title=\textbf{Setting~5. Amount=34}]
Climate Change: Aerosol Radiative Effects, Climate Change: Agriculture, Climate Change: Agriculture \& Pollution, Climate Change: Atmospheric Processes, Climate Change: Biodiversity Loss, Climate Change: Carbon Cycle, Climate Change: Circular Economy, Climate Change: Climate Teleconnections, Climate Change: Corporate Responsibility, Climate Change: Ecosystem Impacts, Climate Change: Energy Demand Management, Climate Change: Energy Efficiency, Climate Change: Energy Sector, Climate Change: Environmental Monitoring, Climate Change: Extreme Weather, Climate Change: Geoengineering, Climate Change: Human Well-being, Climate Change: Hydrological Changes, Climate Change: Impact Modeling, Climate Change: Industrial Emissions, Climate Change: Internal Variability, Climate Change: Marine Ecosystems, Climate Change: Mitigation Pathways, Climate Change: Polar Regions, Climate Change: Policy \& Governance, Climate Change: Public Engagement, Climate Change: Radiative Forcing, Climate Change: Risk Management, Climate Change: Scientific Literature, Climate Change: Sectoral Analysis, Climate Change: Sustainable Products, Climate Change: Temperature Trends, Climate Change: Tourism Impacts, Climate Change: Urban Adaptation.
\end{tcolorbox}

\begin{tcolorbox}[colback=gray!5,colframe=black!40,
                  boxrule=0.3pt,arc=2pt,
                  left=4pt,right=4pt,top=4pt,bottom=4pt,
                  breakable,
                  title=\textbf{Setting~6. Amount=21}]
Climate Change: Carbon Accounting, Climate Change: Climate Monitoring, Climate Change: Ecosystem Services, Climate Change: Emergency Preparedness, Climate Change: Energy Systems, Climate Change: Food Systems, Climate Change: Governance, Climate Change: Human Health, Climate Change: Hydrology \& Flooding, Climate Change: Land Use, Climate Change: Physical Mechanisms, Climate Change: Planetary Geology, Climate Change: Pollution Management, Climate Change: Renewable Energy, Climate Change: Climate Finance, Climate Change: Human Habitability, Climate Change: Infrastructure Development, Climate Change: Misinformation, Climate Change: Ozone Depletion, Climate Change: Paleoclimate, Climate Change: Urbanization.
\end{tcolorbox}

\newlength{\ColA}\setlength{\ColA}{0.13\textwidth} % Category
\newlength{\ColB}\setlength{\ColB}{0.20\textwidth} % Topic
\newlength{\ColC}\setlength{\ColC}{0.67\textwidth} % Mapping

\setlength{\tabcolsep}{3pt}
\renewcommand{\arraystretch}{1.15}

\newcommand{\CategoryCell}[2]{% #1=text, #2=rowspan
  \multirow[t]{#2}{\ColA}{\RaggedRight #1}%
}

% -------------------------------------------------------------
\begin{table*}[htbp]
  \centering
  \begin{adjustbox}{width=\textwidth, max totalheight=\textheight}
  % \small
  \footnotesize
  \begin{tabular}{%
    >{\RaggedRight\arraybackslash}p{\ColA}
    >{\RaggedRight\arraybackslash}p{\ColB}
    >{\RaggedRight\arraybackslash}p{\ColC}
  }
  \toprule
  \textbf{Category} & \textbf{Topic} & \textbf{Mapping to Taxonomy} \\
  \midrule

\CategoryCell{Climate Science Foundations \& Methods}{6} & Atmospheric Science \& Climate Processes & Atmospheric Science (S1, S2); Cloud Processes (S2); Radiative Forcing (S2, S4, S5); Atmospheric Processes (S3, S5); Climate Science (S4); Internal Variability (S5) \\
\cmidrule(lr){2-3}
 & Greenhouse Gas \& Biogeochemical Cycles & Biogeochemical Cycles (S1); Greenhouse Gas Emissions (S1); Fossil Fuels / Fossil Fuel Extraction (S2, S4); Carbon Cycle (S4, S5); Carbon Accounting (S6) \\
\cmidrule(lr){2-3}
 & Oceans, Cryosphere \& Sea-Level Change & Cryosphere Dynamics (S1); Oceanography (S3); Cryosphere Changes (S3, S4); Marine Ecology/Ecosystems (S4, S5); Polar Regions (S5) \\
\cmidrule(lr){2-3}
 & Extreme Weather Events & Extreme Weather Events (S1); Extreme Weather (S2, S3, S4, S5) \\
\cmidrule(lr){2-3}
 & Climate Modeling & Climate Modeling (S1, S2, S3, S4); Impact Modeling (S6) \\
\cmidrule(lr){2-3}
 & Environmental Monitoring & Environmental Monitoring (S3, S5); Data \& Observation (S4); Climate Monitoring (S6) \\
\midrule

\CategoryCell{Ecological Impacts}{3} & Biodiversity Loss & Biodiversity Conservation (S1, S4); Biodiversity Loss (S3, S5); Evolutionary Biology (S3) \\
\cmidrule(lr){2-3}
 & Terrestrial \& Freshwater Ecosystem Changes & Hydrological Change/Changes (S1, S5); Hydrological Impacts (S4); Ecosystem Impacts (S5); Hydrology \& Flooding (S6) \\
\cmidrule(lr){2-3}
 & Marine \& Coastal Ecosystem Changes & Environmental Impact (S1); Oceanography (S3); Marine Ecology/Ecosystems (S4, S5); Ecosystem Impacts (S5) \\
\midrule

\CategoryCell{Human Systems \& Socioeconomic Impacts}{6} & Agriculture \& Food Security & Agriculture (S1, S2, S3, S4, S5); Agricultural Biotechnology (S3); Food Systems (S6) \\
\cmidrule(lr){2-3}
 & Water Resources \& Hydrological Impacts & Hydrological Change/Changes (S1, S5); Water Resources (S3); Hydrological Impacts (S4); Hydrology \& Flooding (S6) \\
\cmidrule(lr){2-3}
 & Human Health \& Well-being & Health Impacts (S1); Public Health (S2, S4); Human Health and Well-being (S3); Human Well-being (S5); Human Health (S6); Human Habitability (S6) \\
\cmidrule(lr){2-3}
 & Social Equity, Vulnerability \& Migration & Social Vulnerability (S1); Societal Transformation (S1); Social Equity (S4); Human-Environment Interaction (S4) \\
\cmidrule(lr){2-3}
 & Urban Systems \& Infrastructure Impacts & Urban Planning (S2, S3); Urban Heat Mitigation (S4); Infrastructure Development (S6); Urbanization (S6) \\
\cmidrule(lr){2-3}
 & Service \& Industry Sector Impacts & Economic Impacts (S1); Tourism Impacts (S2, S3, S5); Economic Assessment (S2); Economic Structures (S2); Sectoral Impacts (S4); Industry Sustainability (S4); Electrification Impacts (S4); Sectoral Analysis (S5) \\
\midrule

\CategoryCell{Adaptation Strategies}{5} & Agricultural \& Food System Adaptation & Adaptation Strategies (S1, S3); Food Systems (S6) \\
\cmidrule(lr){2-3}
 & Urban Planning, Adaptation \& Resilience & Adaptation Strategies (S1, S3); Urban Planning (S2,S3); Urban Heat Mitigation (S4); Urban Adaptation (S5); Infrastructure Development (S6) \\
\cmidrule(lr){2-3}
 & Public Health Adaptation & Health Impacts (S1); Public Health (S2, S4); Human Health and Well-being (S3); Human Well-being (S5); Human Health (S6) \\
\cmidrule(lr){2-3}
 & Public Awareness, Communication \& Community Engagement & Public Perception (S1); Community Engagement (S2, S3); Public Awareness (S3); Public Engagement (S5); Misinformation (S6) \\
\cmidrule(lr){2-3}
 & Natural Resource Management \& Conservation & Adaptation Strategies (S1, S3); Natural Disturbances (S4); Ecosystem Services (S6); Land Use (S6); Pollution Management (S6) \\
\midrule

\CategoryCell{Mitigation Mechanisms}{5} & Climate Policy, Governance \& Finance Mechanism & Policy \& Governance (S1, S3, S4, S5); Governance (S2, S6); Regional Policy (S2); Climate Finance (S3, S6); Financial Risk (S3); Socioeconomic Policy (S4); ESG Investing (S4) \\
\cmidrule(lr){2-3}
 & Energy Transition & Energy Transition (S1, S3); Renewable Energy (S1, S6); Energy Systems (S2, S6); Electrification Impacts (S4); Energy Sector (S5); Energy Efficiency (S5) \\
\cmidrule(lr){2-3}
 & Corporate \& Industry Climate Action & Corporate Sustainability (S1, S3); Sustainable Development (S3, S4); Industry Sustainability (S4); Corporate Responsibility (S5); Sustainable Products (S5); Circular Economy (S5) \\
\cmidrule(lr){2-3}
 & Land Use \& Ecosystem-based Mitigation & Carbon Cycle (S4, S5); Natural Disturbances (S4); Land Use (S6); Ecosystem Services (S6) \\
\cmidrule(lr){2-3}
 & Transport \& Building Emissions Reduction & Transportation (S2, S4); Transportation Emissions (S4); Electrification Impacts (S4); Infrastructure Development (S6) \\
\bottomrule
\end{tabular}
\end{adjustbox}
\caption{Primary mapping from collected topics to the final taxonomy.}
\label{tab:topic_mapping}
\end{table*}

\subsection{Framework Analysis Perspectives}
Under the \textit{Proactive Knowledge Behaviors Framework}, we identify three meaningful comparative perspectives based on the knowledge behaviors among the three actors.

The first perspective is termed \textit{multi-actor behavioral comparison on one target}, which examines behavioral differences among different actors (e.g., different human knowledge providers) when interacting with the same target (e.g., an AI knowledge provider).

The second perspective is termed \textit{intra-actor behavioral comparison}, which focuses on how the same actor exhibits different behaviors when interacting with different targets.

The third perspective is termed \textit{inter-actor behavioral comparison}, which compares behavioral differences across different types of actors.

Based on these three comparative perspectives, with moderate extensions, it becomes possible to systematically operationalize the analyses corresponding to Steps 1--4 in Table~\ref{tab:step-mapping}. We exclude comparisons that are sequential in nature and lack informational value, such as comparing \textit{human ask human} vs.\ \textit{human guide AI}, or \textit{human inform human} vs.\ \textit{human ask AI}. By leveraging these three analytical perspectives, this approach also enables the framework to be systematically extended beyond the context of climate change to other socio-scientific research domains.

\begin{table*}[b]
\centering
\begin{adjustbox}{width=1\textwidth}
\small
\begin{tabular}{
  >{\RaggedRight\arraybackslash}p{0.15\linewidth}  % RQ
  >{\RaggedRight\arraybackslash}p{0.26\linewidth}  % Actors
  >{\RaggedRight\arraybackslash}p{0.30\linewidth}  % Perspective
  >{\RaggedRight\arraybackslash}p{0.34\linewidth}  % Example
}
\toprule
\textbf{Step} & \textbf{Actor(s) Involved} & \textbf{Analytical Perspective} & \textbf{Behavioral Comparison} \\
\midrule

(1) Misalignment & Human Knowledge Seeker + Human Knowledge Provider
& Multi-actor behavioral on one target comparison 
& Seeker Ask LLM vs.\ Provider Guide LLM \\

(2) Similarity & Human Knowledge Seeker 
& Intra-actor behavioral comparison 
& Ask Human vs.\ Ask LLM \\

(3) Reference & Human Knowledge Seeker $\leftrightarrow$ Human Knowledge Provider 
& Inter-actor behavioral comparison 
& Seeker Ask Human vs.\ Provider Inform Human \\

(4) Insights & Human Knowledge Provider 
& Intra-actor behavioral comparison 
& Inform Human vs.\ Guide LLM \\
\bottomrule
\end{tabular}
\end{adjustbox}
\caption{Mapping of analysis steps to actors, analytical perspectives, and behaviors under the \textit{Proactive Knowledge Behaviors Framework}.}
\label{tab:step-mapping}
\end{table*}

\subsection{Dataset and Annotation Examples}
\label{app:examples}
Here we present examples from the eight core datasets used in this study, along with their annotated labels for topic and question type.

\subsubsection{WildChat}

\begin{tcolorbox}[colback=gray!5,colframe=black!40,
                  boxrule=0.3pt,arc=2pt,
                  left=4pt,right=4pt,top=4pt,bottom=4pt,
                  breakable]
{
"text": "Water management is a critical adaptation strategy for climate change. Which of the following does not support this position A. Technology should focus on use of marine water resources B. Finding alternative water stores C. Implementing legislation to ensure the 'fair' distribution of water D. Increasing water conservation in periods of water surplus",\\
"Final\_Topics": ["C2. Water Resources \& Hydrological Impacts", "D5. Natural Resource Management \& Conservation"],\\
"Final\_Question\_Types": {\\
"Intent": ["INTENT\_1a. Fact Lookup"],\\
"Form": ["FORM\_7a. Multiple Choice"]
}
}
\end{tcolorbox}

\begin{tcolorbox}[colback=gray!5,colframe=black!40,
                  boxrule=0.3pt,arc=2pt,
                  left=4pt,right=4pt,top=4pt,bottom=4pt,
                  breakable]
{
"text": "\textit{(Translated from Chinese into English)} 2. The attached file scrippsm.txt is a list of 180 numbers representing atmospheric carbon dioxide concentration, measured in ppv (parts per volume), i.e., the volume of CO2 contained in one million volumes of air. The data are records from the Mauna Loa volcano from January 1996 to December 2010, and this portion of the data was obtained from the Scripps study. (a) Perform a least-squares fit to the CO2 data using the model $f(t)=c_1+c_2t+c_3\cos(2\pi t)+c_4\sin(2\pi t)$, where t is measured in years. Report the optimal fitting coefficients Ci and the RMSE fitting error. Plot the continuous curve from January 1989 to the end of this year, together with the 180 data points. (b) Use your model to predict the CO2 concentrations in May 2004, September 2004, May 2005, and September 2005. These months tend to contain the annual maximum and minimum of the CO2 cycle. The actual recorded values are 380.63, 374.06, 382.45, and 376.73 ppv, respectively. Report the errors at these four points. (c) Add the extra term c5cos(4*pi*t) and redo (a) and (b). Compare the new RMSE and the model errors at the four points. (d) Repeat (c) using the additional term $c_6 t^2$. Which term provides the greatest improvement in the model, (c) or (d)? (e) Add both terms from (c) and (d), and redo (a) and (b). Prepare a table summarizing your results for each part of this problem, and provide an explanation for the outcomes as much as possible.",
"Final\_Topics": ["A5. Climate Modeling", "A2. Greenhouse Gas \& Biogeochemical Cycles"],
"Final\_Question\_Types": {
"Intent": ["INTENT\_2b. Data Analysis / Calculation"],
"Form": ["FORM\_2b. Detailed Multi-paragraph", "FORM\_4a. Tabular Data"]
}
}
\end{tcolorbox}

\begin{tcolorbox}[colback=gray!5,colframe=black!40,
                  boxrule=0.3pt,arc=2pt,
                  left=4pt,right=4pt,top=4pt,bottom=4pt,
                  breakable]
{
"text": "\textit{(Translated from Russian into English)} Greening the economy and economizing ecology: list the methods and examples of using these methods in the interaction between society and nature.",
"Final\_Topics": ["E1. Climate Policy, Governance \& Finance Mechanism"],
"Final\_Question\_Types": {
"Intent": ["INTENT\_1a. Fact Lookup", "INTENT\_3a. General Advice"],
"Form": ["FORM\_3a. Item List"]
}
}
\end{tcolorbox}

\subsubsection{LMSYS-Chat-1M}

\begin{tcolorbox}[colback=gray!5,colframe=black!40,
                  boxrule=0.3pt,arc=2pt,
                  left=4pt,right=4pt,top=4pt,bottom=4pt,
                  breakable]
                  {
"text": "Describe the evolution of succulent plants with reference to past climate and climate change.",\\
"Final\_Topics": ["B2. Terrestrial \& Freshwater Ecosystem Changes"],\\
"Final\_Question\_Types": {\\
"Intent": ["INTENT\_2a. Reasoning / Causal Analysis", "INTENT\_1b. Concept Definition"],\\
"Form": ["FORM\_2b. Detailed Multi-paragraph"]
}
}
\end{tcolorbox}

\begin{tcolorbox}[colback=gray!5,colframe=black!40,
                  boxrule=0.3pt,arc=2pt,
                  left=4pt,right=4pt,top=4pt,bottom=4pt,
                  breakable]
{
"text": "peux tu résume l'article \"Methods to Optimize Carbon Footprint of Buildings in Regenerative Architectural Design with the Use of Machine Learning, publié par Convolutional Neural Network, and Parametric Design NAME\_1 en 2020?",\\
"Final\_Topics": ["E5. Transport \& Building Emissions Reduction"],\\
"Final\_Question\_Types": {\\
"Intent": ["INTENT\_4c. Summarisation"],\\
"Form": ["FORM\_2a. Concise Paragraph"]
}
}
\end{tcolorbox}

\begin{tcolorbox}[colback=gray!5,colframe=black!40,
                  boxrule=0.3pt,arc=2pt,
                  left=4pt,right=4pt,top=4pt,bottom=4pt,
                  breakable]
{
"text": "Task: Analyzing Perspectives on Climate Change Background: Climate change is a complex and multifaceted issue that has been the subject of intense scientific study and public debate. The following paragraph provides some information on the causes and impacts of climate change: Paragraph: Climate change is caused by human activities such as burning fossil fuels, deforestation, and industrial processes, which release greenhouse gases into the atmosphere. These greenhouse gases trap heat from the sun, causing the Earth's temperature to rise and leading to a range of impacts, including more frequent and severe heatwaves, droughts, floods, and storms. Instructions: 1. Read the paragraph above. 2. Create a claim that either expresses skepticism or belief about climate change, using the information presented in the paragraph. 3. For each claim, classify it as either representing the views of a climate skeptic or a climate believer. Use the following format for your claims: Claim format: {Claim statement}. Classification: {Skepticism or Not-skepticism}. You can create as many claims as you like, but make sure you have at least one claim from each class.",\\
"Final\_Topics": ["A2. Greenhouse Gas \& Biogeochemical Cycles", "A4. Extreme Weather Events"],\\
"Final\_Question\_Types": {\\
"Intent": ["INTENT\_5a. General Text"],\\
"Form": ["FORM\_4b. JSON"]
}
}
\end{tcolorbox}

\subsubsection{ClimateQ\&A}

\begin{tcolorbox}[colback=gray!5,colframe=black!40,
                  boxrule=0.3pt,arc=2pt,
                  left=4pt,right=4pt,top=4pt,bottom=4pt,
                  breakable]
{
"text": "can nuclear be a sustainable energy despite the difficulty of managing radioactive waste?",\\
"Final\_Topics": ["E2. Energy Transition"],\\
"Final\_Question\_Types": {\\
"Intent": ["INTENT\_2a. Reasoning / Causal Analysis"],\\
"Form": ["FORM\_2a. Concise Paragraph"]
}
}
\end{tcolorbox}

\begin{tcolorbox}[colback=gray!5,colframe=black!40,
                  boxrule=0.3pt,arc=2pt,
                  left=4pt,right=4pt,top=4pt,bottom=4pt,
                  breakable]
{
"text": "what is the problem with biowaste management",\\
"Final\_Topics": ["A2. Greenhouse Gas \& Biogeochemical Cycles", "E1. Climate Policy, Governance \& Finance Mechanism"],\\
"Final\_Question\_Types": {\\
"Intent": ["INTENT\_2a. Reasoning / Causal Analysis",\\
"INTENT\_1b. Concept Definition"], "Form": ["FORM\_2a. Concise Paragraph", "FORM\_3a. Item List"]
}
}
\end{tcolorbox}

\begin{tcolorbox}[colback=gray!5,colframe=black!40,
                  boxrule=0.3pt,arc=2pt,
                  left=4pt,right=4pt,top=4pt,bottom=4pt,
                  breakable]
{
"text": "What is the Global Warming Potential of R-454C?",\\
"Final\_Topics": ["A2. Greenhouse Gas \& Biogeochemical Cycles"],\\
"Final\_Question\_Types": {\\
"Intent": ["INTENT\_1a. Fact Lookup"],\\
"Form": ["FORM\_1a. Concise Value(s) / Entity(ies)"]
}
}
\end{tcolorbox}

\subsubsection{ClimaQA-Gold}

\begin{tcolorbox}[colback=gray!5,colframe=black!40,
                  boxrule=0.3pt,arc=2pt,
                  left=4pt,right=4pt,top=4pt,bottom=4pt,
                  breakable]
{
"text": "How did the CLAW Hypothesis proposed by Charlson et al. in 1987 suggest a mechanism to stabilize Earth's temperature?",\\
"Final\_Topics": ["A1. Atmospheric Science \& Climate Processes", "A2. Greenhouse Gas \& Biogeochemical Cycles"],\\
"Final\_Question\_Types": {\\
"Intent": ["INTENT\_2a. Reasoning / Causal Analysis", "INTENT\_1b. Concept Definition"],\\
"Form": ["FORM\_1b. Brief Statement"]
}
}
\end{tcolorbox}

\begin{tcolorbox}[colback=gray!5,colframe=black!40,
                  boxrule=0.3pt,arc=2pt,
                  left=4pt,right=4pt,top=4pt,bottom=4pt,
                  breakable]
{
"text": "Biogenic <MASK> contribute to the production of extremely low volatile organic compounds with atmospheric implications.",\\
"Final\_Topics": ["A1. Atmospheric Science \& Climate Processes"],\\
"Final\_Question\_Types": {\\
"Intent": ["INTENT\_1a. Fact Lookup"],\\
"Form": ["FORM\_1a. Concise Value(s) / Entity(ies)"]
}
}
\end{tcolorbox}

\begin{tcolorbox}[colback=gray!5,colframe=black!40,
                  boxrule=0.3pt,arc=2pt,
                  left=4pt,right=4pt,top=4pt,bottom=4pt,
                  breakable]
{
"text": "How does aerosol alter the radiative balance in the atmosphere in a hypothetical scenario where there is a significant increase in anthropogenic aerosol emissions?",\\
"Final\_Topics": ["A1. Atmospheric Science \& Climate Processes"],\\
"Final\_Question\_Types": {\\
"Intent": ["INTENT\_1a. Fact Lookup"],\\
"Form": ["FORM\_7a. Multiple Choice"]
}
}
\end{tcolorbox}

\subsubsection{Clima-Silver}

\begin{tcolorbox}[colback=gray!5,colframe=black!40,
                  boxrule=0.3pt,arc=2pt,
                  left=4pt,right=4pt,top=4pt,bottom=4pt,
                  breakable]
{
"text": "How would the overall number of cloud droplets be affected if the maximum supersaturation in a cloud formation process was lower than usual?",\\
"Final\_Topics": ["A1. Atmospheric Science \& Climate Processes"],\\
"Final\_Question\_Types": {\\
"Intent": ["INTENT\_2a. Reasoning / Causal Analysis"],\\
"Form": ["FORM\_1b. Brief Statement"]
}
}
\end{tcolorbox}

\begin{tcolorbox}[colback=gray!5,colframe=black!40,
                  boxrule=0.3pt,arc=2pt,
                  left=4pt,right=4pt,top=4pt,bottom=4pt,
                  breakable]
{
"text": "Disjunctive <MASK> allows for the assessment of probabilities of exceeding specific thresholds, enabling quantitative evaluation of the risk of inaction in environmental management scenarios.",\\
"Final\_Topics": ["A5. Climate Modeling"],\\
"Final\_Question\_Types": {\\
"Intent": ["INTENT\_1a. Fact Lookup"],\\
"Form": ["FORM\_1a. Concise Value(s) / Entity(ies)"]
}
}
\end{tcolorbox}

\begin{tcolorbox}[colback=gray!5,colframe=black!40,
                  boxrule=0.3pt,arc=2pt,
                  left=4pt,right=4pt,top=4pt,bottom=4pt,
                  breakable]
{
"text": "How does the net greenhouse effect of Arctic low clouds contribute to the Earth's energy budget?",\\
"Final\_Topics": ["A1. Atmospheric Science \& Climate Processes"],\\
"Final\_Question\_Types": {\\
"Intent": ["INTENT\_1a. Fact Lookup"],\\
"Form": ["FORM\_7a. Multiple Choice"]
}
}
\end{tcolorbox}

\subsubsection{Reddit}

\begin{tcolorbox}[colback=gray!5,colframe=black!40,
                  boxrule=0.3pt,arc=2pt,
                  left=4pt,right=4pt,top=4pt,bottom=4pt,
                  breakable]
{
"text": "Former climate change denier here. This has been the hottest winter I’ve experienced (in Texas). Is this a result of climate change or a coincidence? Not being sarcastic. I’ve noticed that, at least in the part of Texas I’m in right now, it’s been in the 80s for most of winter. It got below freezing today, but is expecting to go back into the 80s later this week. I know there’s a difference between weather and climate, but I’ve never seen anything this erratic and weird. Y’all know more than me, so I’m interested in your thoughts.",\\
"Final\_Topics": ["A4. Extreme Weather Events"],\\
"Final\_Question\_Types": {\\
"Intent": ["INTENT\_2a. Reasoning / Causal Analysis", "INTENT\_1c. Clarification / Verification"],\\
"Form": ["FORM\_2a. Concise Paragraph", "FORM\_3a. Item List"]
}
}
\end{tcolorbox}

\begin{tcolorbox}[colback=gray!5,colframe=black!40,
                  boxrule=0.3pt,arc=2pt,
                  left=4pt,right=4pt,top=4pt,bottom=4pt,
                  breakable]
{
"text": "When do you think climate change will become so undeniable that even the most stubborn deniers will no longer be able to ignore its impact? The question is asking when climate change will get so bad that even the biggest skeptics can't deny it anymore. It points to a tipping point where extreme weather, higher temperatures, and obvious signs of damage will make denial impossible. It reflects frustration with the ongoing doubt and the hope that clear proof will finally push everyone to take it seriously and act.",\\
"Final\_Topics": ["A4. Extreme Weather Events", "D4. Public Awareness, Communication \& Community Engagement"],\\
"Final\_Question\_Types": {\\
"Intent": ["INTENT\_2a. Reasoning / Causal Analysis"],\\
"Form": ["FORM\_2b. Detailed Multi-paragraph", "FORM\_2a. Concise Paragraph"]
}
}
\end{tcolorbox}

\begin{tcolorbox}[colback=gray!5,colframe=black!40,
                  boxrule=0.3pt,arc=2pt,
                  left=4pt,right=4pt,top=4pt,bottom=4pt,
                  breakable]
{
"text": "Rising Temperatures, Rapid Aging: How Ambient Heat Affects DNA in Seniors Global warming is driving more frequent extreme heat events, posing severe health risks to senior citizens. By 2050, over 100 million Americans could be affected, with rising hospital admissions and cardiovascular issues. A recent study links heat exposure to epigenetic aging, with research showing DNA methylation changes in heart tissue and immune systems.

What are your thoughts on this study?

Read more here : [Rising Temperatures and Rapid Aging](https://medtigo.com/news/rising-temperatures-rapid-aging-how-ambient-heat-affects-dna-in-seniors/)",\\
"Final\_Topics": ["C3. Human Health \& Well-being", "A4. Extreme Weather Events"],\\
"Final\_Question\_Types": {\\
"Intent": ["INTENT\_2c. Evaluation / Review"],\\
"Form": ["FORM\_2a. Concise Paragraph"]
}
}
\end{tcolorbox}

\subsubsection{SciDCC}

\begin{tcolorbox}[colback=gray!5,colframe=black!40,
                  boxrule=0.3pt,arc=2pt,
                  left=4pt,right=4pt,top=4pt,bottom=4pt,
                  breakable]
{
"text": "Climate. Warmer springs mean more offspring for prothonotary warblers. Climate change contributes to gradually warming Aprils in southern Illinois, and at least one migratory bird species, the prothonotary warbler, is taking advantage of the heat. A new study analyzing 20 years of data found that the warblers start their egg-laying in southern Illinois significantly earlier in warmer springs. This increases the chances that the birds can raise two broods of offspring during the nesting season, researchers found. They report their findings in the journal. Warmer springs in temperate regions of the planet can create a mismatch between when food is available to breeding birds and when their energy demands are highest -- when they are feeding nestlings, said Illinois Natural History Survey avian ecologist Jeffrey Hoover, who conducted the study with INHS principal scientist Wendy Schelsky. If climate change diminishes insect populations at critical moments in the birds' nesting season, food shortages may cause some chicks and even adults to die. Other long-distance migratory birds suffer ill consequences from warmer springs on their breeding grounds. But this mismatch seems not to occur in prothonotary warblers, the researchers found. In the Cache River watershed, where the study was conducted, the warblers nest in swamps and forested wetlands, which are abundant sources of insects throughout the spring and summer. The warblers eat caterpillars, flies, midges, spiders, mayflies and dragonflies. As long as it does not dry up, the wetland habitat provides a steady supply of food to sustain the birds. We studied populations of prothonotary warblers nesting in nest boxes in these areas from 1994–2013, Schelsky said. The researchers captured and color-banded the birds, studying 2,017 nesting female warblers in all. They visited the nests every three to five days from mid-April to early August, keeping track of when each bird laid its first egg and the birds' overall reproductive output for each nesting season. They also compared local April temperature trends with those of the Annual Global Land Temperature Anomaly data compiled by NOAA. They found that local conditions reflected global temperature changes. When local temperatures were warmer in April, the birds tended to lay their first eggs earlier in the spring and produced a greater number of offspring over the course of the breeding season. As the warming trend continues, conditions may change in ways that harm the birds' reproductive capacity. Warmer temperatures could cause local wetlands to dry up during the nesting season, cutting off the steady supply of insects the birds rely on to raise their young and increasing the exposure of nests to predators.",\\
"Final\_Topics": ["B2. Terrestrial \& Freshwater Ecosystem Changes", "B1. Biodiversity Loss"],\\
"Final\_Question\_Types": {\\
"Intent": ["INTENT\_4e. Information Extraction"],\\
"Form": ["FORM\_4b. JSON"]
}
}
\end{tcolorbox}

\subsubsection{IPCC AR6}

\begin{tcolorbox}[colback=gray!5,colframe=black!40,
                  boxrule=0.3pt,arc=2pt,
                  left=4pt,right=4pt,top=4pt,bottom=4pt,
                  breakable]
{
"text": "CMIP5 models with a model top within the stratosphere seriously underestimate the amplitude of the variability of the wintertime NAM expression in the stratosphere, in contrast to CMIP5 models which extend well above the stratopause (Lee and Black, 2015). However, even in the latter models, the stratospheric NAM events and their downward influence on the troposphere are insufficiently persistent (Charlton-Perez et al., 2013; Lee and Black, 2015). Increased vertical resolution does not show any significant added value in reproducing the structure and magnitude of the tropospheric NAM (Lee and Black, 2013) nor in the NAO predictability assessed in a seasonal prediction context with a multi-model approach (Butler et al., 2016). On the other hand, there is mounting evidence that a correct representation of the Quasi Biennial Oscillation, extratropical stratospheric dynamics (the polar vortex and sudden stratospheric warmings), and related troposphere-stratosphere coupling, as well as their interplay with ENSO, are important for NAO/NAM timing (Scaife et al., 2016; Karpechko et al., 2017; Domeisen, 2019; Domeisen et al., 2019), in spite of underestimated troposphere–stratosphere coupling found in models compared to observations (O’Reilly et al., 2019b).",\\
"Final\_Topics": ["A5. Climate Modeling", "A1. Atmospheric Science \& Climate Processes"],\\
"Final\_Question\_Types": {\\
"Intent": ["INTENT\_9z. Others"],\\
"Form": ["FORM\_9z. Others"]
}
}
\end{tcolorbox}

\subsection{LLM Knowledge Requirements by User Intent}

We introduce how each user intent maps to the knowledge dimensions:
F=Factual, C=Conceptual, P=Procedural, M=Metacognitive.
We mark \textbf{$\star$} for primary (indispensable) knowledge and $\circ$ for auxiliary (helpful but not core). See Table~\ref{tab:K_r_1}, \ref{tab:K_r_2}, \ref{tab:K_r_3}, \ref{tab:K_r_4}, \ref{tab:K_r_5}, \ref{tab:K_r_6}, \ref{tab:K_r_7}, \ref{tab:K_r_8}.
\label{app: knowledge_taxonomy_reason}

\newcolumntype{Y}{>{\RaggedRight\arraybackslash}X}
% ---------- INTENT_1 ----------
% \subsection{INTENT\_1 (Information retrieval)}
\begin{table*}[htbp]
\centering
\begin{adjustbox}{max width=\linewidth}
\begin{tabularx}{\linewidth}{lccccY}
\toprule
\textbf{Topic} & \textbf{F} & \textbf{C} & \textbf{P} & \textbf{M} & \textbf{Reason} \\
\midrule
1a-Fact lookup & \textbf{$\star$} &  &  & $\circ$ & Retrieve specific facts/data $\rightarrow$ atomic verifiable information; occasional uncertainty disclosure. \\
1b-Concept definition & \textbf{$\star$} & \textbf{$\star$} &  &  & Definitions/characteristics $\rightarrow$ boundaries and attributes of a concept. \\
1c-Clarification / verification & \textbf{$\star$} & $\circ$ &  & \textbf{$\star$} & Verify truth/resolve ambiguity $\rightarrow$ fact checking plus criteria selection/uncertainty handling. \\
\bottomrule
\end{tabularx}
\end{adjustbox}
\caption{Knowledge Requirements for INTENT\_1 (Information retrieval).}
\label{tab:K_r_1}
\end{table*}

% ---------- INTENT_2 ----------
% \subsection{INTENT\_2 (Analysis / evaluation)}
\begin{table*}[htbp]
\centering
\begin{adjustbox}{max width=\linewidth}
\begin{tabularx}{\linewidth}{lccccY}
\toprule
\textbf{Topic} & \textbf{F} & \textbf{C} & \textbf{P} & \textbf{M} & \textbf{Reason} \\
\midrule
2a-Reasoning / causal analysis &  & \textbf{$\star$} & $\circ$ & $\circ$ & Explain causes/effects/comparisons $\rightarrow$ conceptual relations frame the analysis; procedural steps and strategy are supportive. \\
2b-Data analysis / calculation &  & $\circ$ & \textbf{$\star$} & $\circ$ & Perform numerical/statistical computation $\rightarrow$ execution of methods and pipelines is core; method choice is supportive. \\
2c-Evaluation / review &  & $\circ$ & $\circ$ & \textbf{$\star$} & Assess and rate $\rightarrow$ establishing criteria, weighting evidence, and judgment are metacognitive. \\
\bottomrule
\end{tabularx}
\end{adjustbox}
\caption{Knowledge Requirements for INTENT\_2 (Analysis / evaluation).}
\label{tab:K_r_2}
\end{table*}

% ---------- INTENT_3 ----------
% \subsection{INTENT\_3 (Guidance / support)}
\begin{table*}[htbp]
\centering
\begin{adjustbox}{max width=\linewidth}
\begin{tabularx}{\linewidth}{lccccY}
\toprule
\textbf{Topic} & \textbf{F} & \textbf{C} & \textbf{P} & \textbf{M} & \textbf{Reason} \\
\midrule
3a-General advice &  & $\circ$ & $\circ$ & \textbf{$\star$} & Non-technical recommendations $\rightarrow$ context-sensitive trade-offs and strategy selection. \\
3b-Technical assistance / troubleshooting &  & $\circ$ & \textbf{$\star$} & $\circ$ & Practical help resolving issues $\rightarrow$ reproducible diagnose–test–verify procedure. \\
3c-Planning / strategy &  & $\circ$ & $\circ$ & \textbf{$\star$} & Plans/roadmaps/schedules $\rightarrow$ goal decomposition, prioritization, and monitoring. \\
3d-Teaching / skill building &  & $\circ$ & \textbf{$\star$} & $\circ$ & Tutorials/step-by-step guidance $\rightarrow$ executable sequences are central. \\
\bottomrule
\end{tabularx}
\end{adjustbox}
\caption{Knowledge Requirements for INTENT\_3 (Guidance / support).}
\label{tab:K_r_3}
\end{table*}

% ---------- INTENT_4 ----------
% \subsection{INTENT\_4 (Text transformation)}
\begin{table*}[htbp]
\centering
\begin{adjustbox}{max width=\linewidth}
\begin{tabularx}{\linewidth}{lccccY}
\toprule
\textbf{Topic} & \textbf{F} & \textbf{C} & \textbf{P} & \textbf{M} & \textbf{Reason} \\
\midrule
4a-Translation &  & \textbf{$\star$} & \textbf{$\star$} &  & Convert between languages $\rightarrow$ semantic/pragmatic mapping (C) executed via translation procedures (P). \\
4b-Rewrite &  & $\circ$ & \textbf{$\star$} & $\circ$ & Paraphrase/change tone $\rightarrow$ controlled transformation process; style/genre knowledge assists. \\
4c-Summarisation &  & \textbf{$\star$} & \textbf{$\star$} &  & Condense to key points $\rightarrow$ discourse structure grasp (C) + filtering/compression steps (P). \\
4d-Format conversion &  &  & \textbf{$\star$} &  & Transform file/data formats $\rightarrow$ structural mapping procedure. \\
4e-Information extraction & $\circ$ & $\circ$ & \textbf{$\star$} &  & Extract entities/facts $\rightarrow$ extraction rules/pipeline; schema awareness can help. \\
\bottomrule
\end{tabularx}
\end{adjustbox}
\caption{Knowledge Requirements for INTENT\_4 (Text transformation).}
\label{tab:K_r_4}
\end{table*}

% ---------- INTENT_5 ----------
% \subsection{INTENT\_5 (Creative / generative)}
\begin{table*}[htbp]
\centering
\begin{adjustbox}{max width=\linewidth}
\begin{tabularx}{\linewidth}{lccccY}
\toprule
\textbf{Topic} & \textbf{F} & \textbf{C} & \textbf{P} & \textbf{M} & \textbf{Reason} \\
\midrule
5a-General text generation &  & \textbf{$\star$} & $\circ$ & $\circ$ & Open-ended writing $\rightarrow$ genre/discourse patterns guide content organization. \\
5b-Creative story / poem / lyrics &  & \textbf{$\star$} & $\circ$ & $\circ$ & Narrative/poetic forms and conventions drive creation. \\
5c-Hypothetical scenario &  & \textbf{$\star$} & $\circ$ & $\circ$ & Counterfactual ``what if'' $\rightarrow$ conceptual modeling of assumptions/relations. \\
5d-Role-play / dialogue simulation &  & \textbf{$\star$} & $\circ$ & $\circ$ & Personas and pragmatics (register, intent) are conceptual; execution is secondary. \\
5e-Multimodal creation &  & \textbf{$\star$} & \textbf{$\star$} & $\circ$ & Prompts for images/audio/video $\rightarrow$ cross-modal semantics (C) plus construction procedures (P). \\
\bottomrule
\end{tabularx}
\end{adjustbox}
\caption{Knowledge Requirements for INTENT\_5 (Creative / generative).}
\label{tab:K_r_5}
\end{table*}

% ---------- INTENT_6 ----------
% \subsection{INTENT\_6 (Practical / structured outputs)}
\begin{table*}[htbp]
\centering
\begin{adjustbox}{max width=\linewidth}
\begin{tabularx}{\linewidth}{lccccY}
\toprule
\textbf{Topic} & \textbf{F} & \textbf{C} & \textbf{P} & \textbf{M} & \textbf{Reason} \\
\midrule
6a-Operational writing &  & $\circ$ & \textbf{$\star$} & $\circ$ & Emails/reports/ads $\rightarrow$ templates, drafting workflow, and deliverable constraints. \\
6b-Code solution & $\circ$ & $\circ$ & \textbf{$\star$} & $\circ$ & Write/fix code $\rightarrow$ syntax, APIs, test–debug procedures; specific constants as facts. \\
6c-Formulas \& expressions & $\circ$ & $\circ$ & \textbf{$\star$} &  & LaTeX/Excel/regex $\rightarrow$ expression/syntax construction procedures. \\
6d-Structured generation &  & $\circ$ & \textbf{$\star$} & $\circ$ & Tables/schedules/checklists $\rightarrow$ schema-constrained generation and formatting steps. \\
\bottomrule
\end{tabularx}
\end{adjustbox}
\caption{Knowledge Requirements for INTENT\_6 (Practical / structured outputs).}
\label{tab:K_r_6}
\end{table*}

% ---------- INTENT_7 ----------
% \subsection{INTENT\_7 (Navigation / access)}
\begin{table*}[htbp]
\centering
\begin{adjustbox}{max width=\linewidth}
\begin{tabularx}{\linewidth}{lccccY}
\toprule
\textbf{Topic} & \textbf{F} & \textbf{C} & \textbf{P} & \textbf{M} & \textbf{Reason} \\
\midrule
7a-Website navigation &  & $\circ$ & \textbf{$\star$} & $\circ$ & Open URLs/pages $\rightarrow$ path/step execution; choose shortest/robust route. \\
7b-System / resource access &  & $\circ$ & \textbf{$\star$} & $\circ$ & Open files/systems $\rightarrow$ operation sequences with basic safety/permission strategy. \\
\bottomrule
\end{tabularx}
\end{adjustbox}
\caption{Knowledge Requirements for INTENT\_7 (Navigation / access).}
\label{tab:K_r_7}
\end{table*}

% ---------- INTENT_8 ----------
% \subsection{INTENT\_8 (Social / engagement)}
\begin{table*}[htbp]
\centering
\begin{adjustbox}{max width=\linewidth}
\begin{tabularx}{\linewidth}{lccccY}
\toprule
\textbf{Topic} & \textbf{F} & \textbf{C} & \textbf{P} & \textbf{M} & \textbf{Reason} \\
\midrule
8a-Greeting / small talk &  & \textbf{$\star$} & $\circ$ & $\circ$ & Casual conversation $\rightarrow$ pragmatics/etiquette schemas; light execution/adjustment. \\
8b-Entertainment / engagement &  & \textbf{$\star$} & $\circ$ & $\circ$ & Jokes/games $\rightarrow$ genre/game conventions; rule execution secondary. \\
8c-Emotional support / empathy &  & $\circ$ & $\circ$ & \textbf{$\star$} & Supportive/empathetic responses $\rightarrow$ boundary setting, validation, escalation judgment. \\
\bottomrule
\end{tabularx}
\end{adjustbox}
\caption{Knowledge Requirements for INTENT\_8 (Social / engagement).}
\label{tab:K_r_8}
\end{table*}

%% file: Sections/Appendix_3.tex
\section{Additional Results}
\label{app:add-results}

\subsection{Overall Distribution}
\label{app:overall_distribution}

\begin{figure*}[h]
    \centering
    \includegraphics[width=0.7\linewidth,page=1]{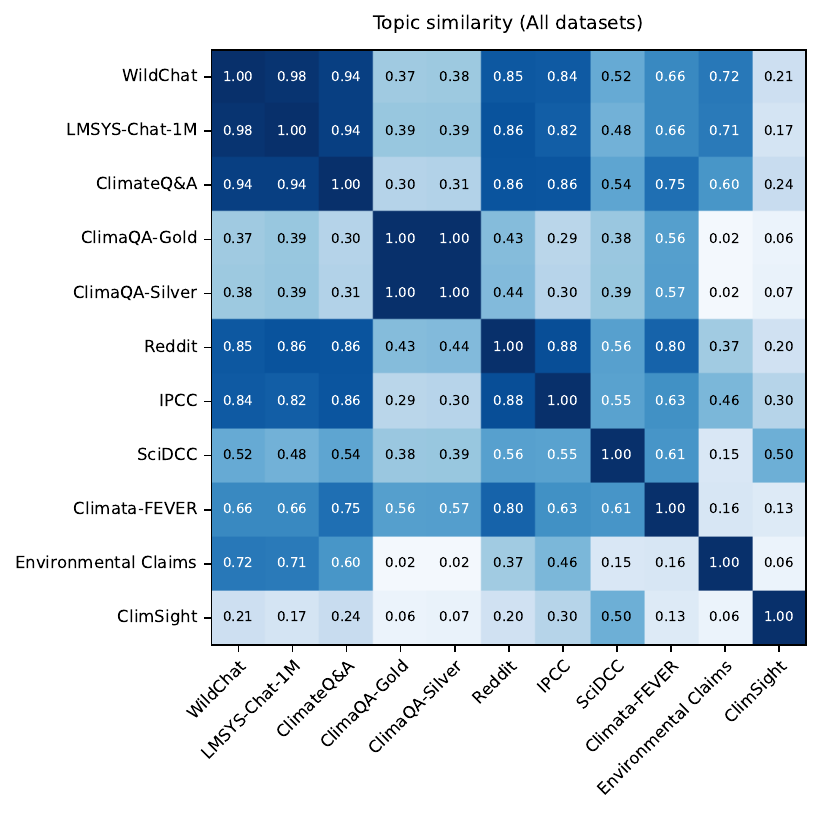}
    \caption{Topic Similarity for All 11 Datasets.}
    \label{fig:af24}
\end{figure*}

\begin{figure*}[h]
    \centering
    \includegraphics[width=0.7\linewidth,page=1]{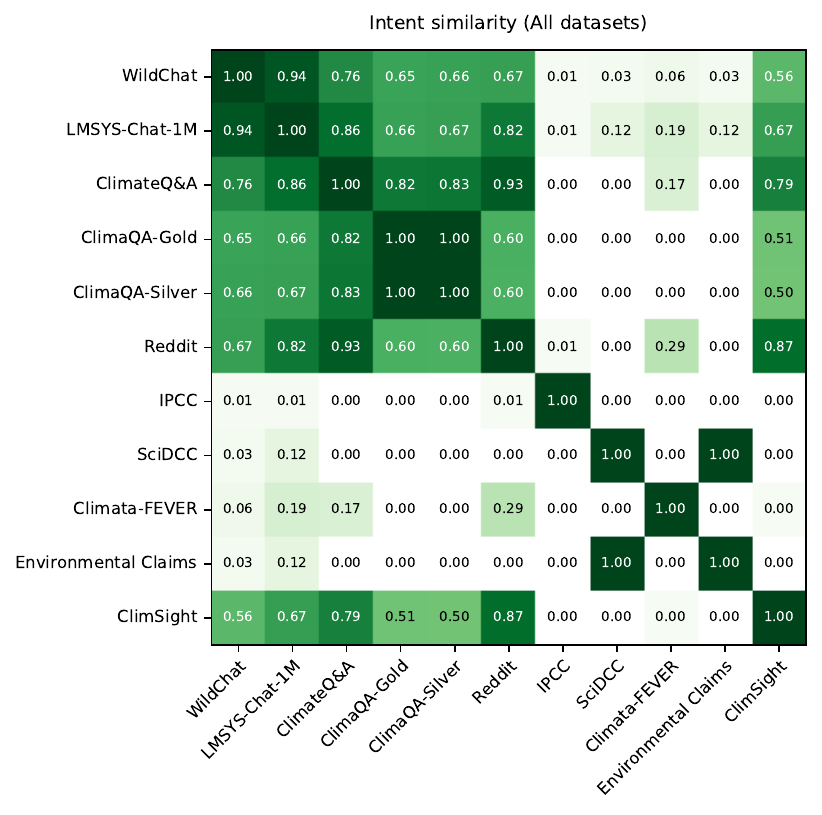}
    \caption{Intent Similarity for All 11 Datasets.}
    \label{fig:af25}
\end{figure*}

\begin{figure*}[h]
    \centering
    \includegraphics[width=0.7\linewidth,page=1]{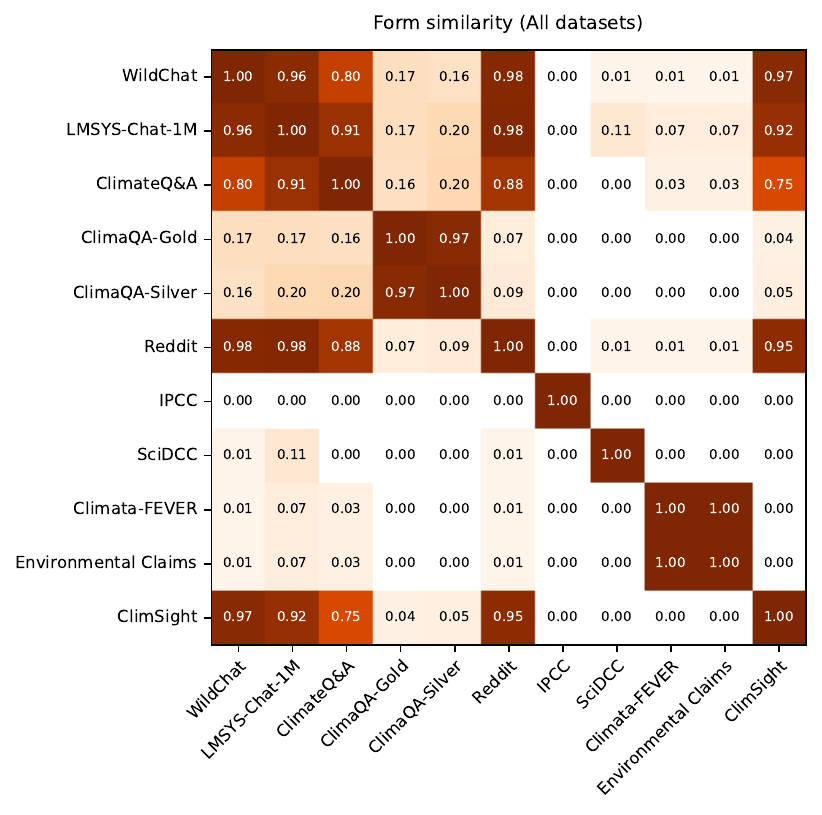}
    \caption{Form Similarity for All 11 Datasets.}
    \label{fig:af26}
\end{figure*}

\begin{figure*}[h]
    \centering
    \includegraphics[width=1.0\linewidth,page=1]{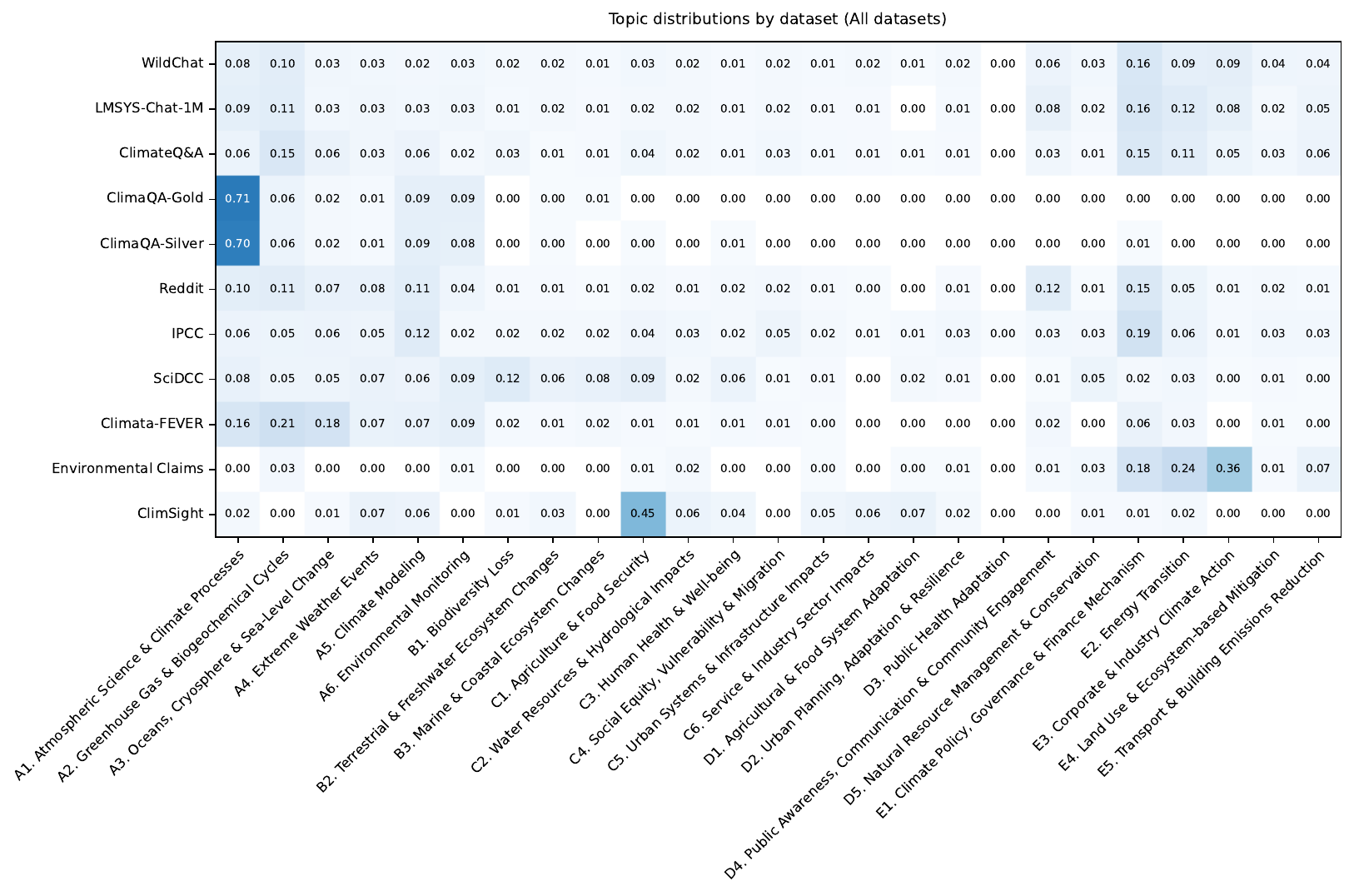}
    \caption{Topic Distribution for All 11 Datasets.}
    \label{fig:af21}
\end{figure*}

\begin{figure*}[h]
    \centering
    \includegraphics[width=1.0\linewidth,page=1]{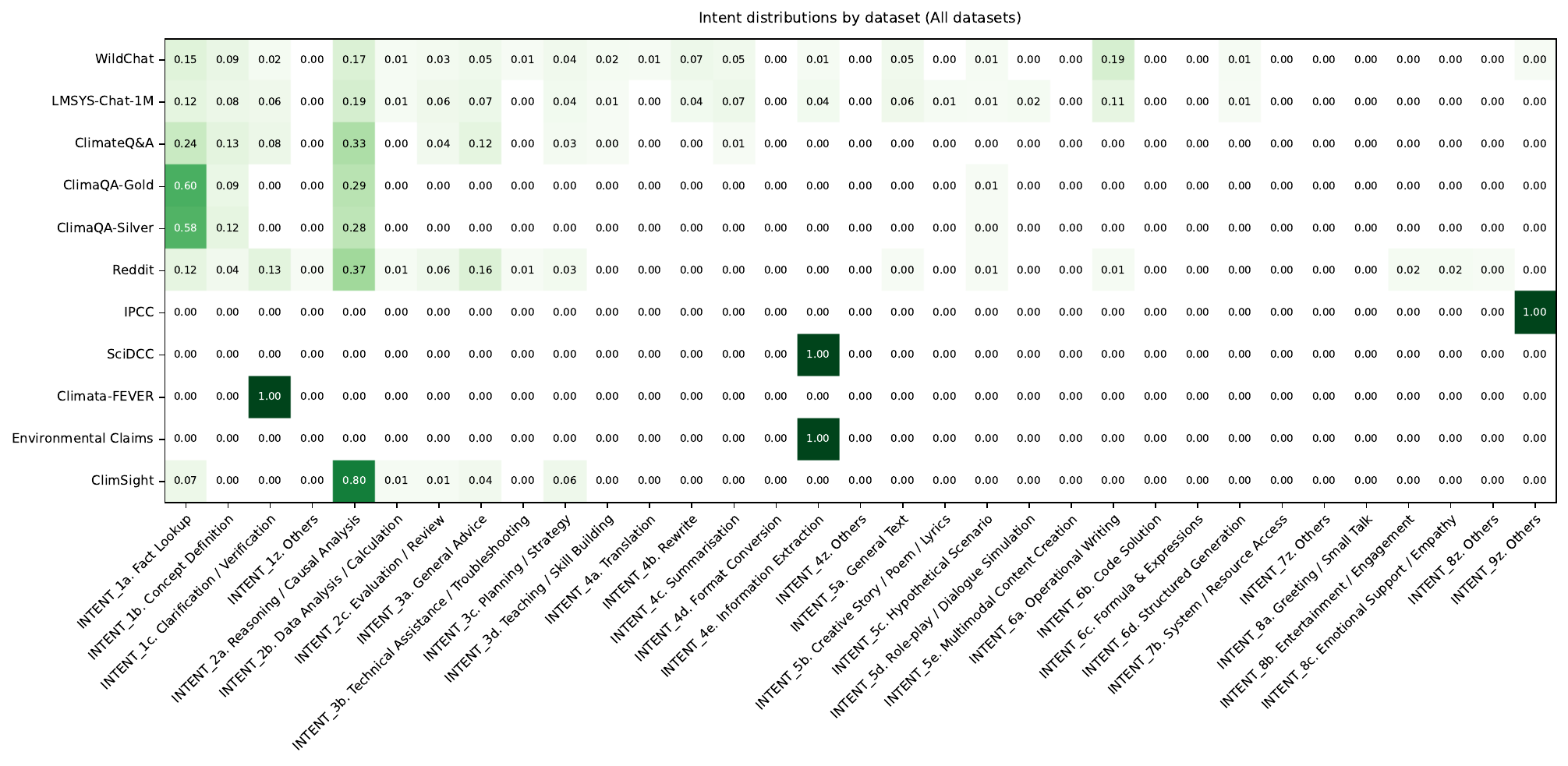}
    \caption{Intent Distribution for All 11 Datasets.}
    \label{fig:af22}
\end{figure*}

\begin{figure*}[h]
    \centering
    \includegraphics[width=0.95\linewidth,page=1]{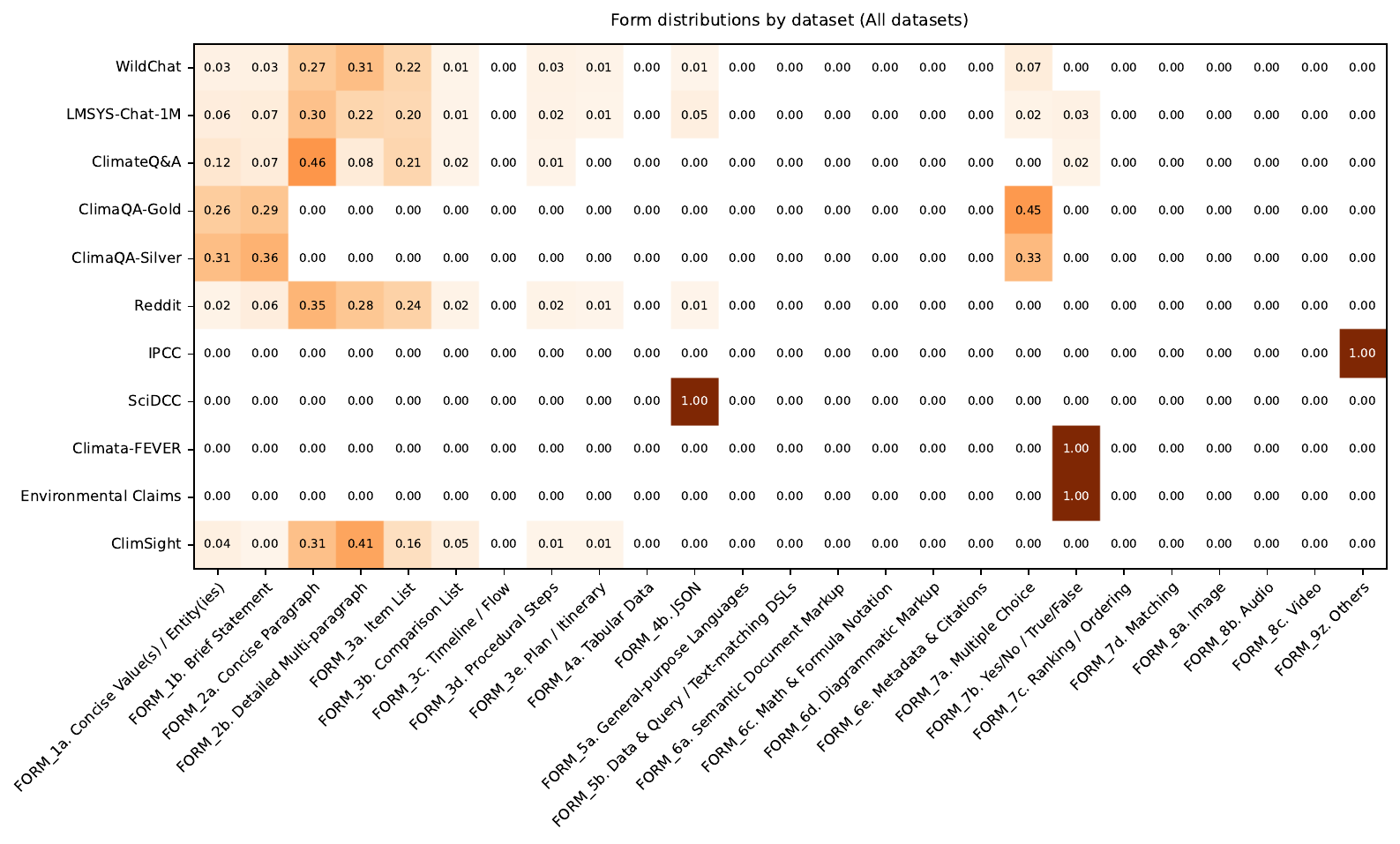}
    \caption{Form Distribution for All 11 Datasets.}
    \label{fig:af23}
\end{figure*}

The complete Topic, Intent, and Form similarities and distributions for all 11 datasets are shown in Figures~\ref{fig:af24}, \ref{fig:af25}, \ref{fig:af26}, \ref{fig:af21}, \ref{fig:af22}, and \ref{fig:af23}.
The Intent labels for \textit{SciDCC}, \textit{Climate-FEVER}, and \textit{Environmental Claims} are fixed according to their original tasks. For \textit{ClimaQA-Gold}, \textit{ClimaQA-Silver}, \textit{SciDCC}, \textit{Climate-FEVER}, and \textit{Environmental Claims}, the Form labels are fixed based on the datasets' original answer types. 
For \textit{IPCC}, both the intent and form are assigned to \textit{9z. Others}.

\subsection{Cross-analysis}
\begin{figure*}[h]
    \centering
    \includegraphics[width=0.8\linewidth,page=1]{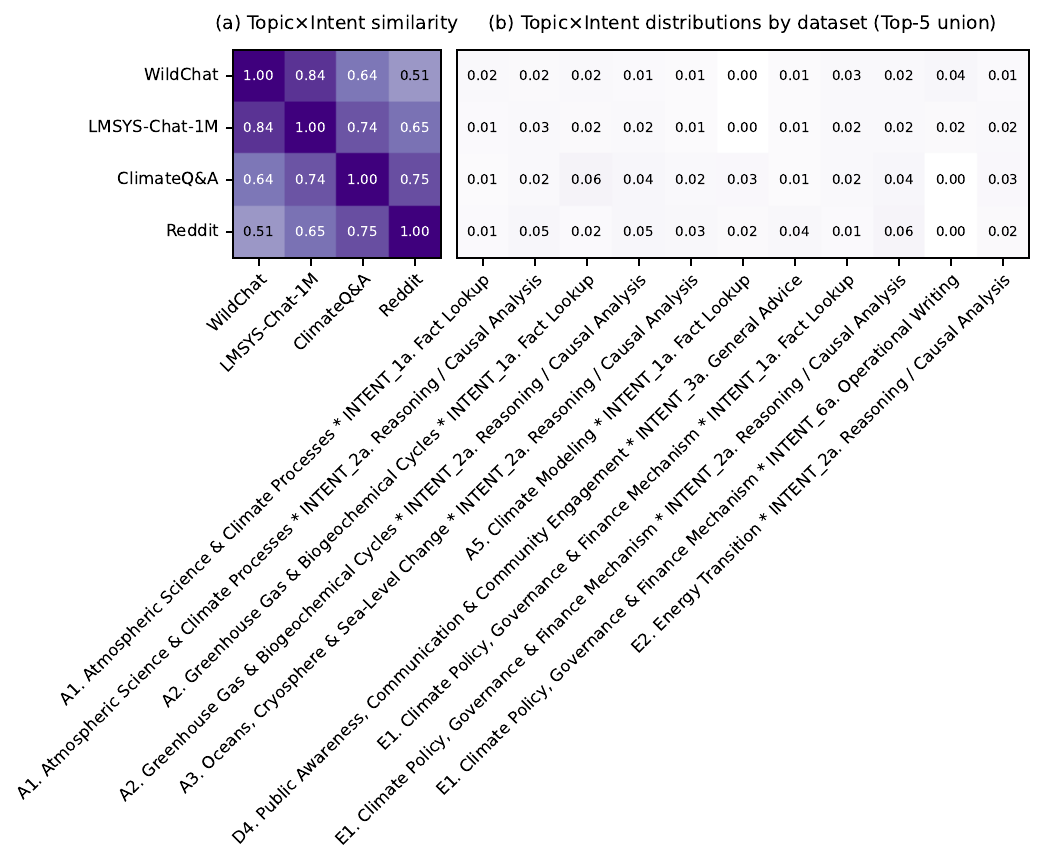}
    \caption{Topic$\times$Intent cross-analysis.}
    \label{fig:af27}
\end{figure*}

\begin{figure*}[h]
    \centering
    \includegraphics[width=0.8\linewidth,page=1]{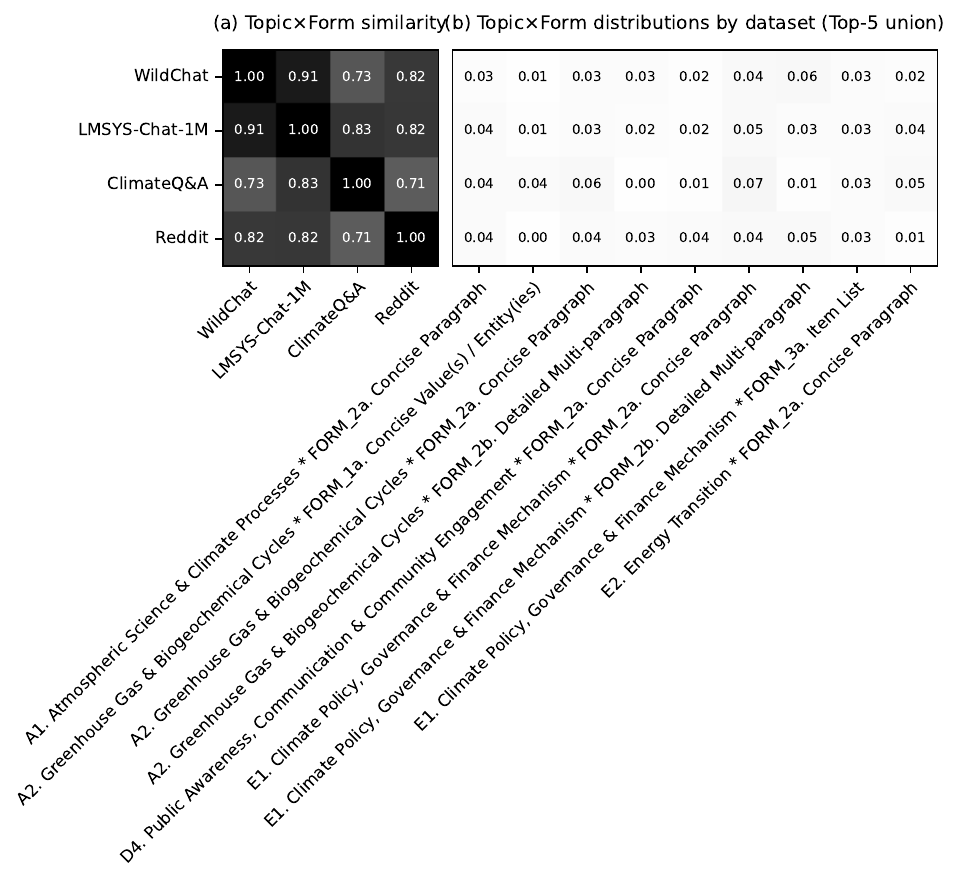}
    \caption{Topic$\times$Form cross-analysis.}
    \label{fig:af28}
\end{figure*}

\begin{figure*}[h]
    \centering
    \includegraphics[width=0.8\linewidth,page=1]{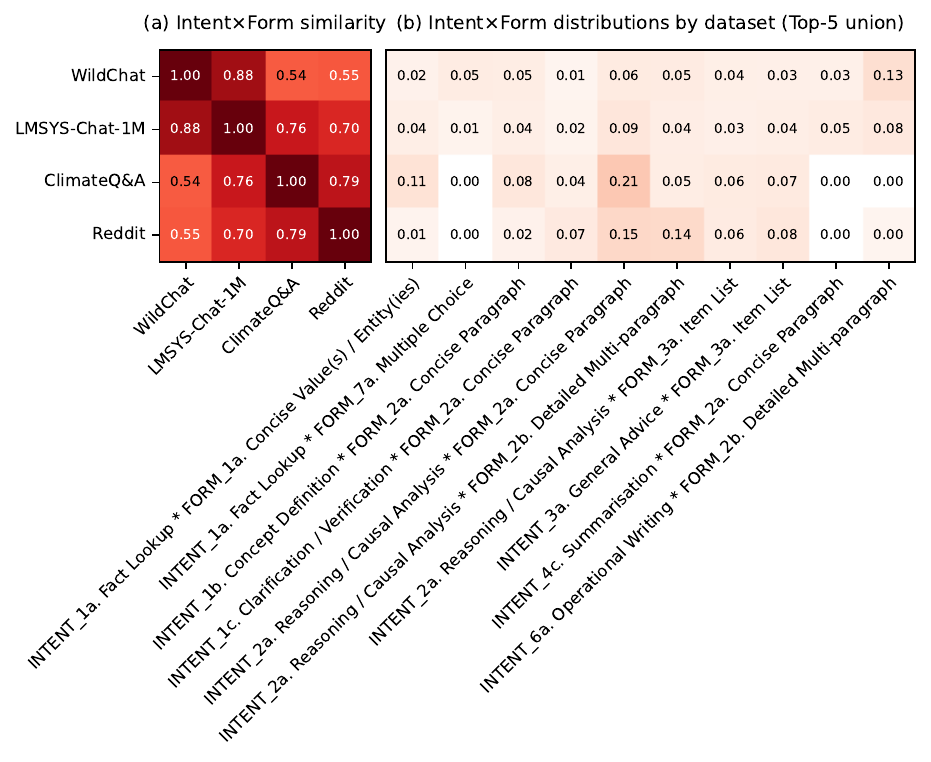}
    \caption{Intent$\times$Form cross-analysis.}
    \label{fig:af29}
\end{figure*}

We further conduct a cross-analysis of the \textit{Topic}, \textit{Intent}, and \textit{Form} of questions posed by humans (including both queries to LLMs and questions directed to other humans). Specifically, we examine Topic$\times$Intent, Topic$\times$Form, and Intent$\times$Form (see Figure~\ref{fig:af27}, ~\ref{fig:af28} ~\ref{fig:af29}). Overall, although Topic$\times$Intent and Topic$\times$Form show relatively high similarity across datasets, their distributions are diffuse, with no strongly dominant pairings. By contrast, the results for Intent$\times$Form align more closely with intuition: on Reddit, ``\textit{INTENT\_2a. Reasoning / Causal Analysis} $\times$ \textit{FORM\_2a/2b}'' accounts for higher shares (about 15\% and 14\%); in ClimateQ\&A, ``\textit{INTENT\_2a} $\times$ \textit{FORM\_2a}'' reaches as high as 21\%.

\subsection{Analysis with Merged Topic Lists}
\label{app:2 merge analysis}
\begin{figure*}[h]
    \centering
    \includegraphics[width=0.7\linewidth,page=1]{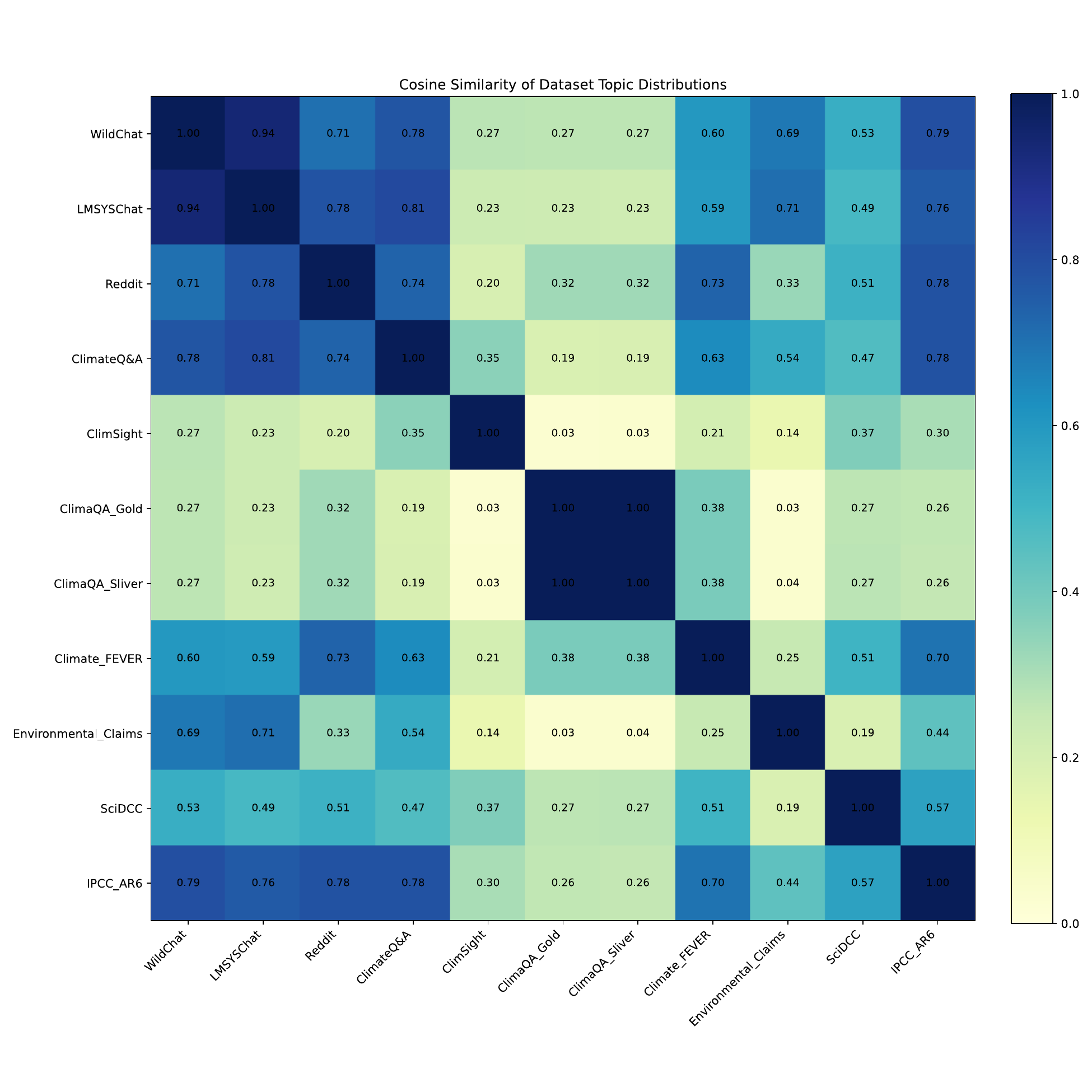}
    \caption{Topic Similarity for All 11 Datasets under S1 Merge Setting.}
    \label{fig:mt1}
\end{figure*}

\begin{figure*}[h]
    \centering
    \includegraphics[width=0.7\linewidth,page=1]{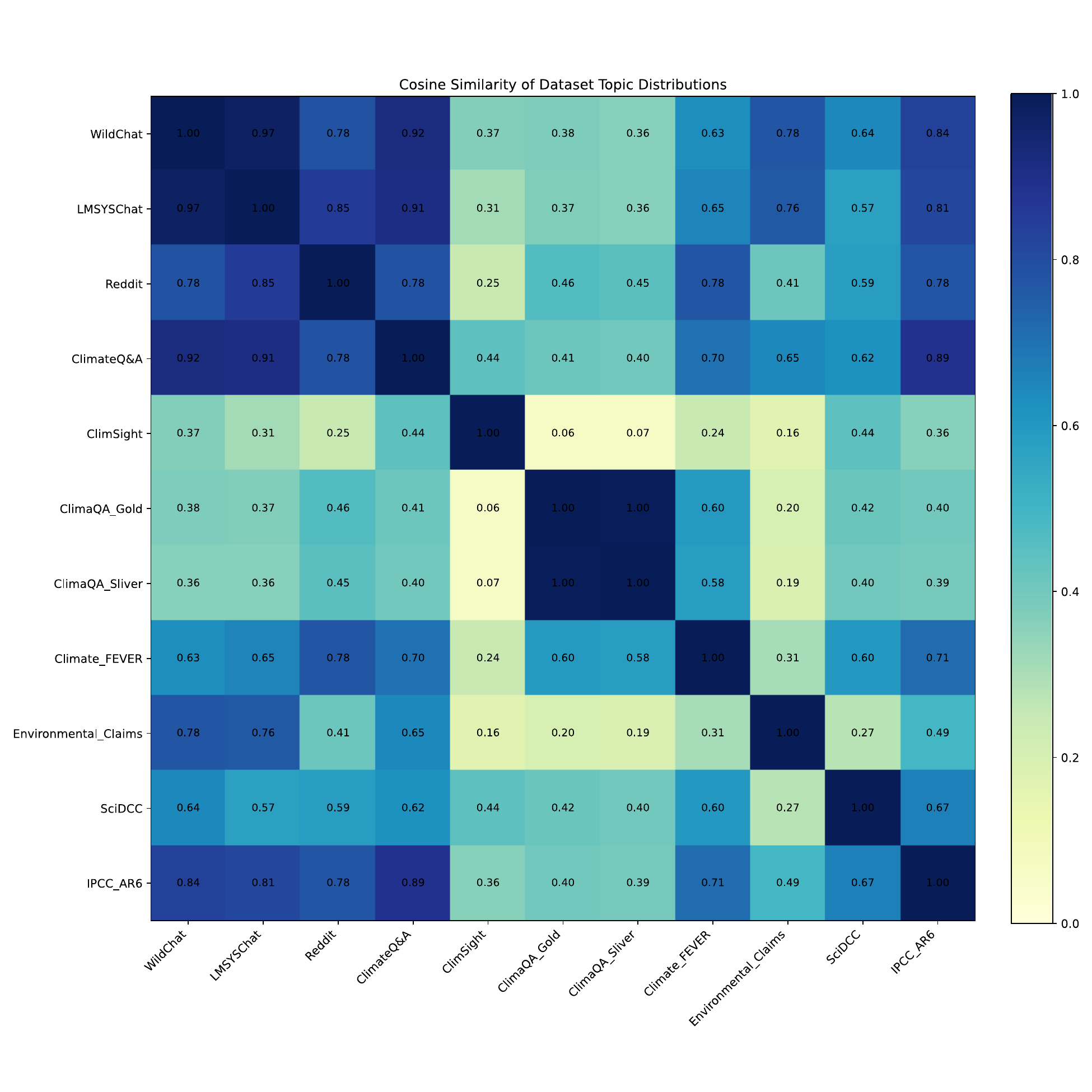}
    \caption{Topic Similarity for All 11 Datasets under S3 Merge Setting.}
    \label{fig:mt3}
\end{figure*}

\begin{figure*}[h]
    \centering
    \includegraphics[width=0.7\linewidth,page=1]{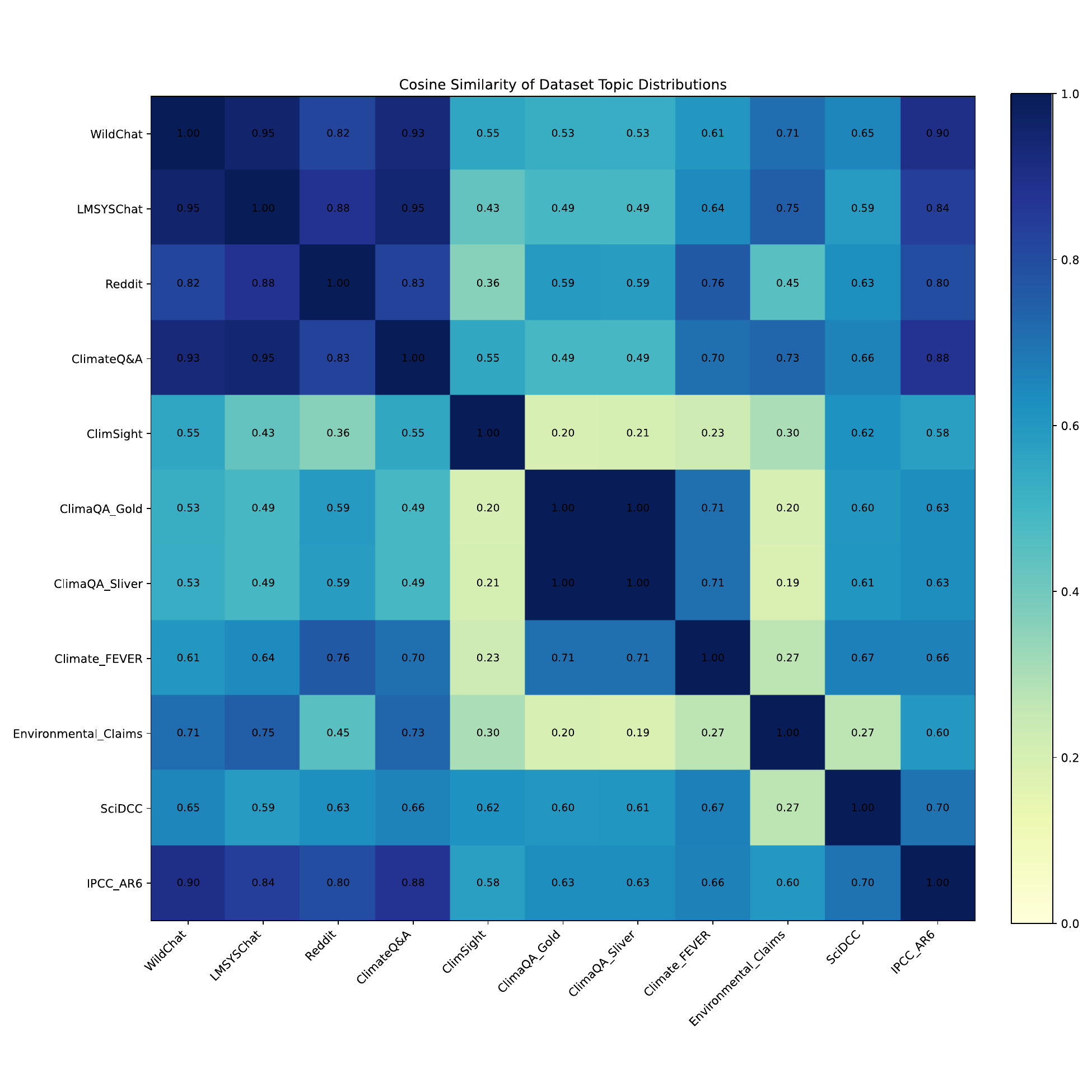}
    \caption{Topic Similarity for All 11 Datasets under S4 Merge Setting.}
    \label{fig:mt4}
\end{figure*}

\begin{figure*}[h]
    \centering
    \includegraphics[width=0.7\linewidth,page=1]{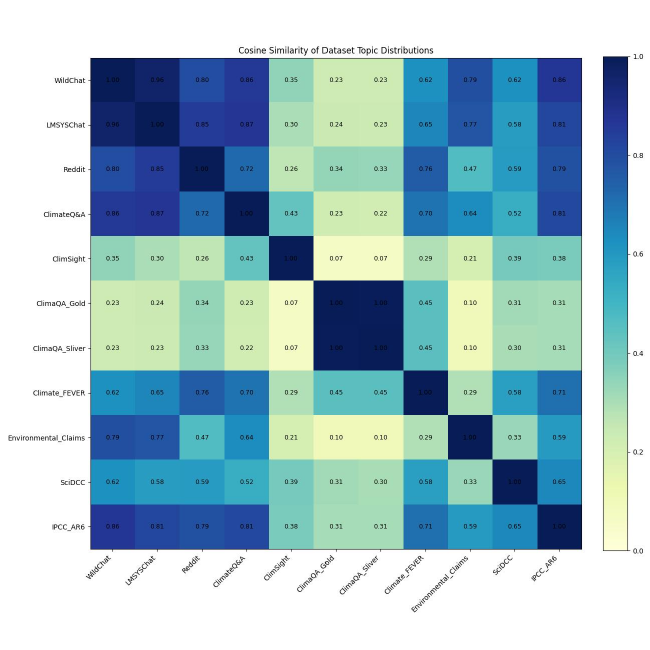}
    \caption{Topic Similarity for All 11 Datasets under S5 Merge Setting.}
    \label{fig:mt5}
\end{figure*}

\begin{figure*}[h]
    \centering
    \includegraphics[width=0.7\linewidth,page=1]{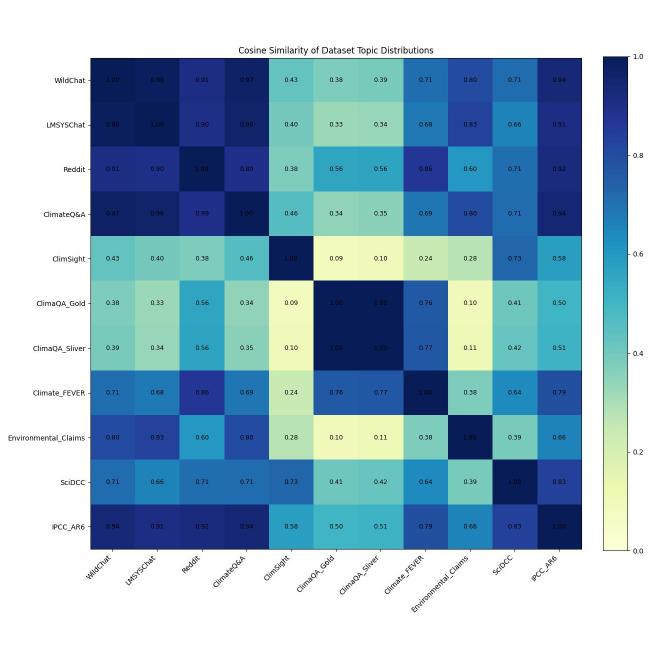}
    \caption{Topic Similarity for All 11 Datasets under S6 Merge Setting.}
    \label{fig:mt6}
\end{figure*}

For completeness, we also examined the topic list obtained directly after the topic merging step (Figure~\ref{fig:mt1},{fig:mt2},~\ref{fig:mt3},~\ref{fig:mt4},~\ref{fig:mt5},~\ref{fig:mt6}). An analysis based on this merged topic list yields results that are broadly consistent with those obtained after applying our taxonomy and performing topic reassignment, this reflects the rationality and effectiveness of our Topic Merging algorithm design. However, to ensure the highest possible accuracy and alignment between topics and the designed classification scheme, we still conducted the topic reassignment procedure before performing the final analysis.

\begin{figure*}[h]
    \centering
    \includegraphics[width=0.7\linewidth,page=1]{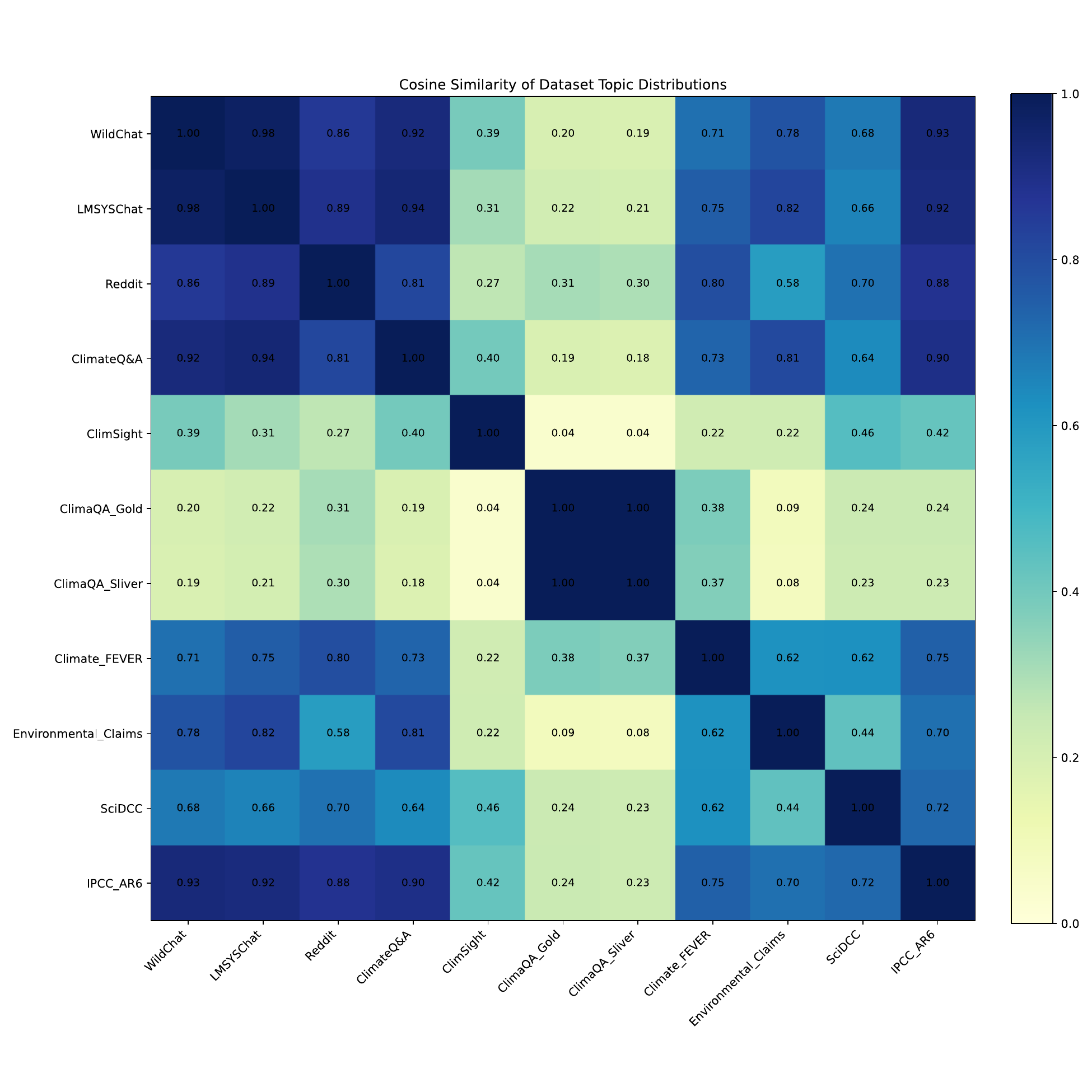}
    \caption{Topic Similarity for All 11 Datasets under S2 Merge Setting.}
    \label{fig:mt2}
\end{figure*}

\subsection{More Visualization}
\label{app:more_visualization}

\begin{figure*}[h]
    \centering
    \includegraphics[width=1.0\linewidth,page=2]{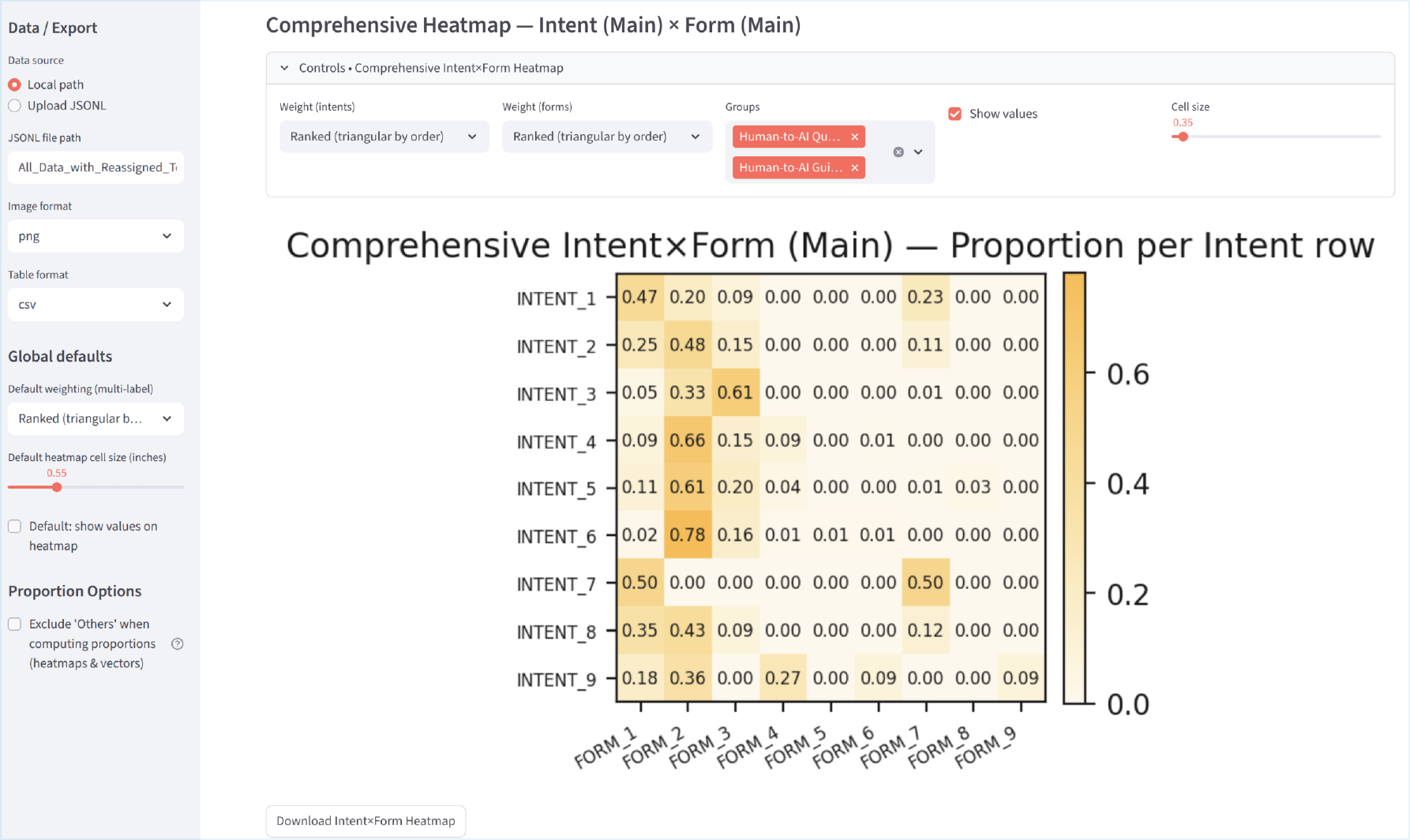}
    \caption{Interactive data visualization enabling flexible data selection, sample weighting, and both multidimensional and cross-dimensional analysis. Shown here is a partial cross-analysis between Intent and Form.}
    \label{fig:af2_visual_web}
\end{figure*}

We built a local, interactive data-visualization web application (Figure~\ref{fig:af2_visual_web}) that supports comprehensive, in-depth, and cross-dimensional analysis across seven analytical dimensions. The web provides visualizations through tables, heatmaps, bar charts, and differential bar charts. It allows analysis by individual datasets or by groups (i.e., datasets under a specific knowledge behavior). Users can choose whether to include “Others” data, apply any of three different data-weighting methods, and conduct analysis at either the Category level or the more granular Topic/Type level. 

Specifically, by visualizing the results under three different weighting methods, we can still arrive at conclusions that are broadly consistent with the main analysis of this paper. The three weighting methods are defined as follows:
\begin{enumerate}
    \item \textbf{Label-count weighting}: each label is assigned an equal weight of $1$, and the total is normalized by the sum of all labels.
    \item \textbf{Per-sample weighting}: for each sample with $K$ labels, the weight is evenly distributed, assigning $\frac{1}{K}$ to each label.
    \item \textbf{Ranked weighting} (Used for ): labels are weighted according to their order using a triangular scheme, where higher-ranked labels receive greater weights. This weighting method is used in the main body of the paper.
\end{enumerate}

\subsection{Topic Merging Tree}
\label{app:topic_merging_tree}

\begin{figure*}[htbp]
    \centering
    \includegraphics[width=1.0\linewidth,page=1]{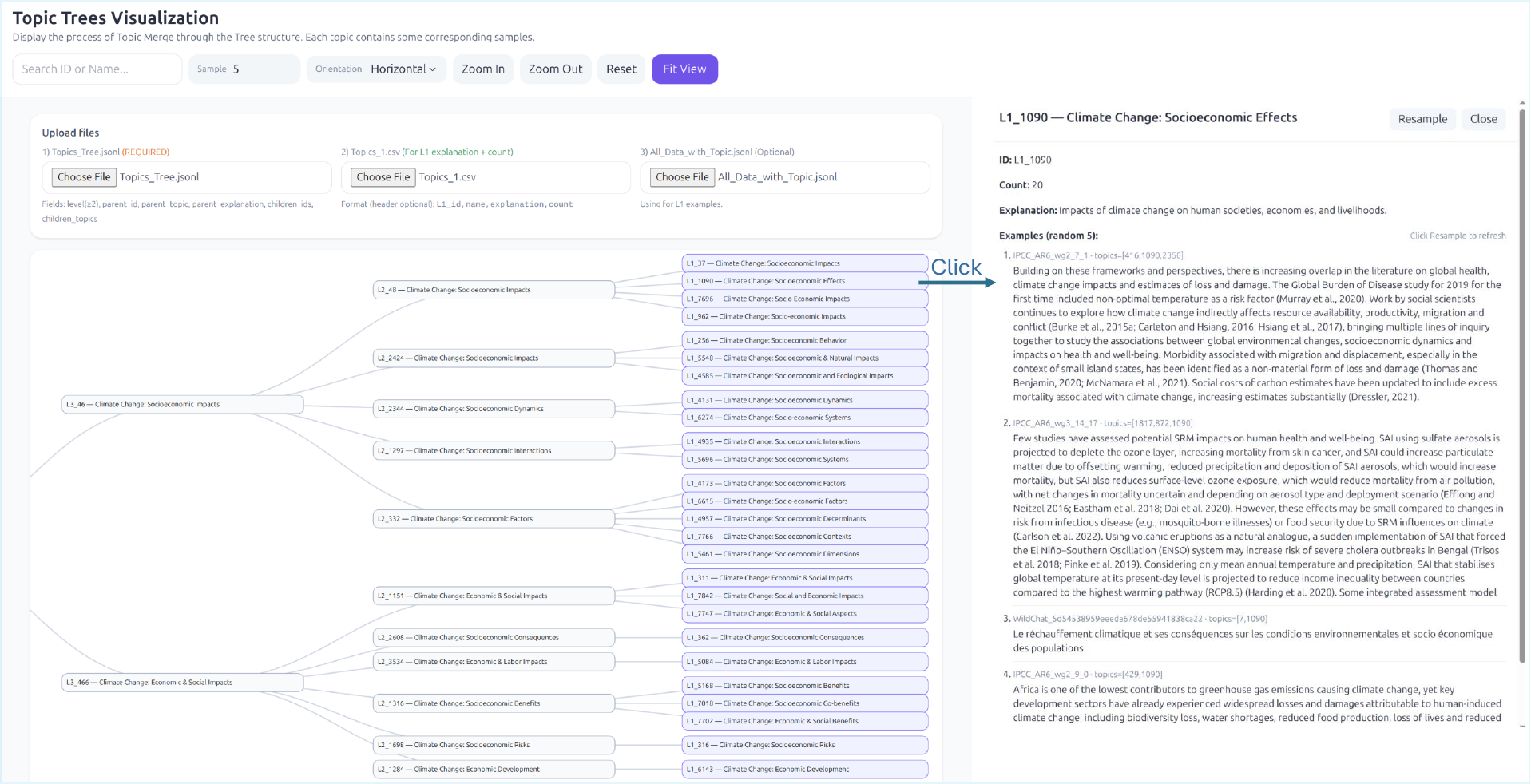}
    \caption{Visualization of the topic-merging process, with interactive access to each topic's explanation, sample count, and randomly sampled instances. Leaf nodes are displayed in purple, while other nodes are shown in gray.}
    \label{fig:af2_tree}
\end{figure*}

We store the Topic Merging process using a tree structure and visualize it in an HTML page (as shown in Figure~\ref{fig:af2_tree}) to make it easier to inspect the merging workflow and its quality, and to support experimental design. Each topic can be clicked to reveal its id, corresponding data volume, explanation, and randomly sampled examples. Localized visualization via search is also supported.

\subsection{Human Verification}
\label{app:human_verification}

\begin{figure*}[h]
    \centering
    \includegraphics[width=1.0\linewidth,page=1]{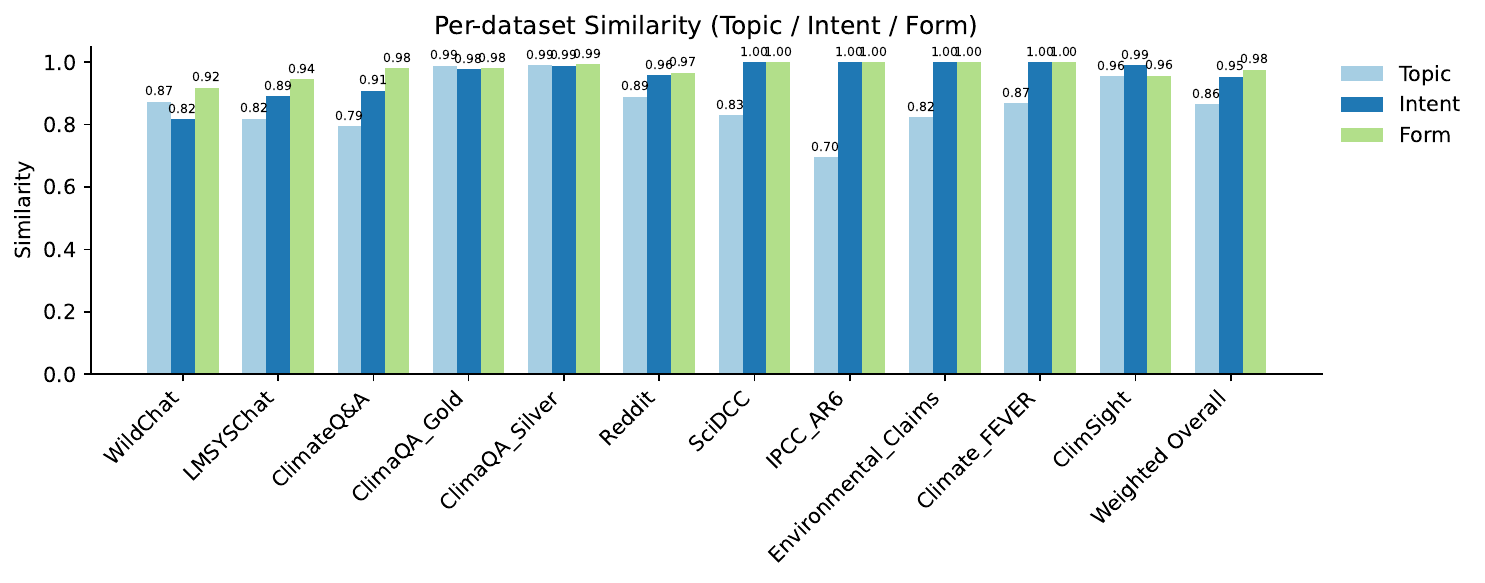}
    \caption{Similarity between the sampled validation subset and the full dataset.}
\label{fig:dataset_similarity}

    \label{fig:verif}
\end{figure*}

We randomly sampled 150 data instances from the full dataset, including 15 instances from each of the eight core datasets and 10 instances from each of the three auxiliary datasets. We mapped the original labels to high-level taxonomy categories (Topic, Intent, and Form) for evaluation. Figure~\ref{fig:dataset_similarity} shows that the sampled validation subset closely matches the full dataset in terms of Topic, Intent, and Form distributions, supporting its representativeness. Six annotators (four PhD-level, one Master’s-level, and one undergraduate) each labeled 50 samples. The instructions for the annotators are as follows:

\begin{tcolorbox}[colback=gray!5,colframe=black!40,
                  boxrule=0.3pt,arc=2pt,
                  left=4pt,right=4pt,top=4pt,bottom=4pt,
                  breakable]
{
Annotate the data with Topic and Question\_Type (Intent + Form). 

\begin{itemize}
\item Each data topic/intent/form can be annotated with 1--3 labels, meaning a single data instance may ultimately have 3--9 labels. 

\item When multiple labels exist, please try to rank them by relevance as much as possible, with the most relevant ones placed first.

\item Some of the intents or forms in the dataset have already been filled in, because those items don’t require intent or form labeling.

\item Within the Topic categories, some type labels under C. Human Systems \& Socioeconomic Impacts and D. Adaptation Strategies appear similar. However, Category C emphasizes impacts, while Category D emphasizes responses and adaptation.

\item In addition, distinguishing between D. Adaptation Strategies and E. Mitigation Mechanisms can also be difficult. The goal of adaptation is to cope with climate change impacts that have already occurred or are unavoidable --- climate change is already happening, and adaptation aims to make social, economic, and ecological systems more capable of resisting or withstanding these shocks. Mitigation, on the other hand, aims to reduce the magnitude of climate change itself --- that is, to lower greenhouse gas emissions or enhance carbon sinks so that global warming becomes less severe.
\end{itemize}

\textbf{Files:}
\begin{itemize}
\item Taxonomy\_Figure.pdf contains two figures of the taxonomy. In it, the topic taxonomy provides several descriptions, and the Topic and Type columns are the dimensions we need to annotate (e.g., A5. Climate Modeling, INTENT\_5a. General Text, FORM\_2a. Concise Paragraph).

\item The Label\_List folder provides the complete sets of labels.

\item The Prompt folder contains descriptions, explanations, clues and examples for determining Topic and Question Type.

\item Examples.jsonl provides several fully annotated examples.
\end{itemize}
}
\end{tcolorbox}

Model--human agreement was measured using Jaccard similarity and Micro-F1, with 95\% confidence intervals (CI) estimated via 1,000-round bootstrap resampling. As shown in Table~\ref{tab:segment_metrics}, the LLM exhibits consistently high agreement with human across all categories (Jaccard scores of 0.706, 0.743, and 0.783 on topic, intent and form, respectively). The relatively narrow CI widths (approximately 0.10--0.12) suggest that the sample size is sufficient to support stable estimation of model--human agreement under the current evaluation setting. We further evaluated human inter-annotator agreement using pairwise Cohen kappa. The average pairwise Cohen kappa scores for Topic, Intent, and Form are 0.594, 0.656, and 0.735, respectively, indicating moderate to substantial agreement.

\begin{table*}[b]
\centering
\begin{adjustbox}{max width=\textwidth}
\scriptsize
\setlength{\tabcolsep}{4pt}
\renewcommand{\arraystretch}{1.15}
\begin{tabular}{
  >{\RaggedRight\arraybackslash}p{0.08\linewidth}  % Model (A, B, Avg)
  >{\RaggedRight\arraybackslash}p{0.10\linewidth}  % Metric (Jaccard, Micro-F1)
  *{15}{c}                                         % scores + CIs
}
\toprule
 & &
 \multicolumn{3}{c}{1--50} &
 \multicolumn{3}{c}{51--100} &
 \multicolumn{3}{c}{101--150} &
 \multicolumn{6}{c}{Overall} \\
\cmidrule(lr){3-5}\cmidrule(lr){6-8}\cmidrule(lr){9-11}\cmidrule(lr){12-17}
\textbf{Annoator} & \textbf{Metric} &
Topic & Intent & Form &
Topic & Intent & Form &
Topic & Intent & Form &
Topic & CI & Intent & CI & Form & CI \\
\midrule
A        & Jaccard   &
0.710 & 0.737 & 0.737 &
0.780 & 0.855 & 0.810 &
0.713 & 0.687 & 0.727 &
0.734 & [$0.671$, $0.789$] &
0.759 & [$0.701$, $0.813$] &
0.757 & [$0.696$, $0.814$] \\
A        & Micro-F1  &
0.739 & 0.739 & 0.713 &
0.787 & 0.864 & 0.821 &
0.735 & 0.712 & 0.707 &
0.755 & [$0.699$, $0.806$] &
0.772 & [$0.716$, $0.822$] &
0.745 & [$0.684$, $0.804$] \\
\midrule
B        & Jaccard   &
0.737 & 0.727 & 0.807 &
0.655 & 0.717 & 0.830 &
0.643 & 0.737 & 0.790 &
0.678 & [$0.620$, $0.743$] &
0.727 & [$0.669$, $0.786$] &
0.809 & [$0.756$, $0.859$] \\
B        & Micro-F1  &
0.754 & 0.739 & 0.796 &
0.688 & 0.719 & 0.844 &
0.632 & 0.737 & 0.811 &
0.691 & [$0.634$, $0.752$] &
0.732 & [$0.674$, $0.789$] &
0.817 & [$0.764$, $0.864$] \\
\midrule
Avg(A,B) & Jaccard   &
0.723 & 0.732 & 0.772 &
0.718 & 0.786 & 0.820 &
0.678 & 0.712 & 0.758 &
0.706 & -- &
0.743 & -- &
0.783 & -- \\
Avg(A,B) & Micro-F1  &
0.747 & 0.739 & 0.755 &
0.737 & 0.792 & 0.832 &
0.684 & 0.724 & 0.759 &
0.723 & -- &
0.752 & -- &
0.781 & -- \\
\midrule
Agr(A,B) & Cohen's $\kappa$ & 0.596 & 0.646 & 0.661 & 0.630 & 0.701 & 0.797 & 0.555 & 0.632 & 0.748 & 0.605 & -- &
0.663 & -- &
0.745 & -- \\
\bottomrule
\end{tabular}
\end{adjustbox}
\caption{Segment-wise and overall performance (mean with 95\% confidence intervals).}
\label{tab:segment_metrics}
\end{table*}